\documentclass{article}

\usepackage[preprint]{corl_2026} 
\usepackage{graphicx}
\usepackage{amsmath}
\usepackage{amssymb}
\usepackage{centernot}
\usepackage{bbm}
\usepackage{booktabs}
\usepackage{graphicx}
\usepackage{multirow}
\usepackage{makecell}
\usepackage{threeparttable}
\usepackage[table]{xcolor}
\usepackage{wrapfig}
\usepackage{caption}
\usepackage{subcaption}

\newcommand{\bestcurrent}[1]{\textcolor{red!75!black}{#1}}
\newcommand{\bestfuture}[1]{\textcolor{orange!85!black}{#1}}
\newcommand{\bestaction}[1]{\textcolor{blue!75!black}{#1}}

\definecolor{jointwam}{HTML}{E8F3FF} 
\definecolor{seqwam}{HTML}{EAF7EA}   
\definecolor{auxwam}{HTML}{FFF3E0}   

\title{Beyond Task Success: Behavioral and Representational Diagnostics for WAM and VLA}

\author{
  Hung PQ.~Mai\\
  National Economics University, Vietnam \\
  N2TP Technology \\
  \And
  Bin Zhu\thanks{Corresponding author} \\
  Singapore Management University \\
  \texttt{binzhu@smu.edu.sg} \\
  \And
  Tuan Do \\
  Phenikaa University, Vietnam \\
  N2TP Technology \\
  \texttt{tuan.do@n2tp.com}
}

\begin{document}
\maketitle

\begin{center}
\includegraphics[width=\textwidth]{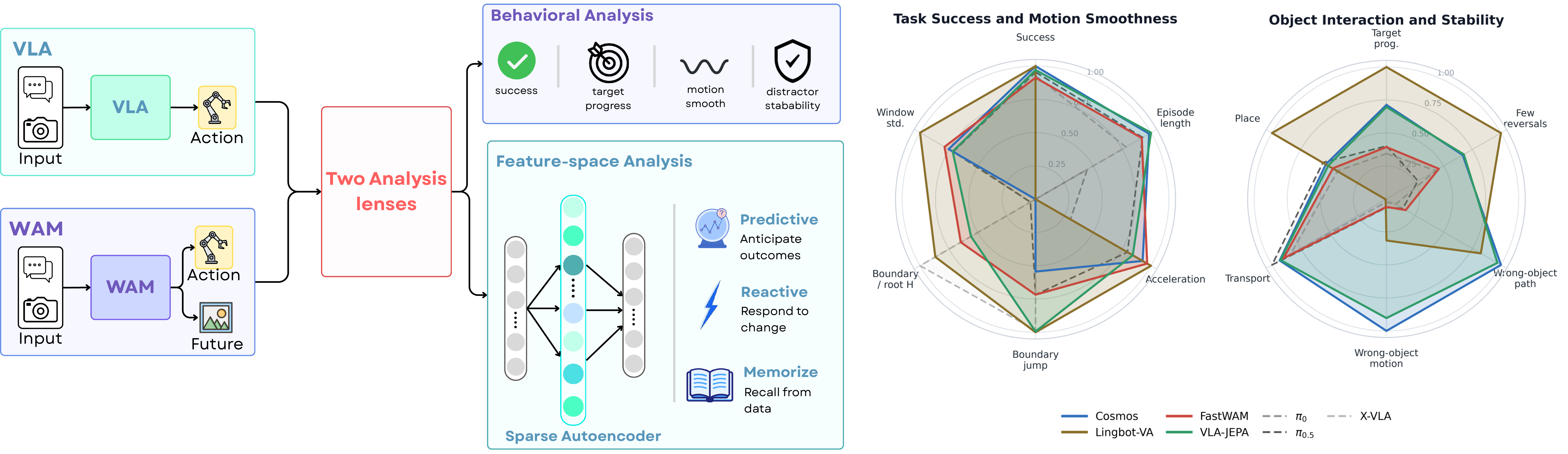}
\captionof{figure}{\small \textbf{Overview of our WAM--VLA analysis framework.}
We compare direct VLA policies and WAM variants through two complementary lenses:
behavioral rollout diagnostics and sparse-autoencoder feature-space analysis.
The framework evaluates not only task success, but also motion smoothness,
target-object progress, distractor stability, and future-oriented internal
representations.}
\label{fig:cover}
\end{center}


\begin{abstract}
Vision-language-action (VLA) policies and World-Action Models (WAM) represent two increasingly important paradigms for robotic manipulation. However, it remains unclear whether future prediction in WAMs leads to behaviorally meaningful improvements beyond final task success. In this paper, we ask whether WAMs merely add future prediction, or whether they change robot behavior and internal representations in ways that are actionable for control. We introduce a model-agnostic diagnostic framework that compares WAMs and VLAs through two complementary lenses: behavioral rollout analysis and sparse-autoencoder-based feature analysis. The behavioral protocol measures action dynamics consistency, target-object progress, distractor disturbance, and runtime cost. The feature-space protocol characterizes internal representations as memorized, reactive, or predictive, revealing whether models encode future-oriented structure. Across LIBERO and RoboTwin2.0, we evaluate 7 policies spanning direct VLAs and joint, sequential, and auxiliary WAMs. Our results show that success alone hides key differences: WAMs often improve object-level behavior and target selectivity, but their gains depend on architecture and incur higher inference cost. Sequential WAMs show the clearest predictive structure, while auxiliary and joint WAMs respectively compress or entangle future information. These findings suggest future directions for WAMs design to preserve behaviorally actionable future representations for efficient manipulation.
\end{abstract}




\section{Introduction}

\vspace{-0.2cm}

Vision-language-action (VLA) policies~\cite{rt2, pi0,pi05,octo_2023}
have become a dominant paradigm for general-purpose robot manipulation. By mapping visual observations and language instructions directly to actions or action chunks, recent VLA models have shown strong performance across diverse manipulation tasks. However, most VLA policies remain fundamentally reactive action predictors: given the current observation and task instruction, they predict what the robot should do next, without explicitly modeling how the scene should evolve as a consequence of those actions. World-Action Models (WAMs) \cite{dreamzero, lingbot-va2026} have recently emerged as a promising alternative. Instead of predicting actions alone, WAMs couple action generation with future prediction. 
Some models jointly generate future observations and actions \cite{cosmospolicy,dreamzero}, some first
imagine future states before acting \cite{lingbot-va2026,gu2026dream4manip}, and others use future prediction only as an auxiliary training signal~\cite{vlajepa2026,yuan2026fastwam}. In principle, this future-oriented modeling should make robot policies less reactive and more coherent: a policy that can anticipate how an object should move, deform, or change state may be better able to select the correct object, avoid distractors, maintain consistent progress, and recover from intermediate uncertainty.

Previous work \cite{zhang2026worldactionmodelsgeneralize} empirically compared the success rates between WAMs and VLAs families. Yet it remains unclear whether WAMs actually provide these benefits beyond improving final task success. A higher success rate does not reveal how a policy succeeds. It does not tell us whether the robot moves smoothly, whether the target object makes steady progress toward the goal, or whether the model internally represents future task evolution in a way that is useful for action selection. Conversely, a WAM may contain explicit future prediction modules but still fail to translate future information into better control. This raises the central question of this paper: \textbf{Do WAMs merely add future prediction, or do they change robot behavior and internal representations in ways that are behaviorally actionable?}

We study this question through a model-agnostic diagnostic framework for comparing WAMs and VLA policies. Instead of treating task success as the sole measure of policy quality, we analyze two complementary aspects. As shown in Fig.~\ref{fig:cover}, first, we perform \textbf{behavioral rollout diagnostics}, measuring action dynamics, chunk-boundary consistency, target-object progress, wrong-object disturbance, manipulation-stage completion, and runtime cost. These metrics reveal whether a policy produces stable, selective, and goal-directed interaction, rather than merely completing the task. Second, we perform \textbf{feature-space diagnostics} using Sparse Autoencoders (SAEs~\cite{huben2024sparse,lan2025sparse}), which allow us to characterize internal features as memorized, reactive, or predictive. This provides a representation-level view of whether WAMs expose future-oriented structure that is largely absent from direct VLA policies.

Across LIBERO and RoboTwin2.0, we evaluate 7 policies spanning direct VLAs and three WAM families: joint/parallel WAMs, sequential WAMs, and auxiliary/training-only WAMs. Our analysis shows that success alone hides important differences. WAMs often improve object-level behavior, including stronger target-object progress and reduced wrong-object disturbance, but their benefits are architecture-dependent. Sequential WAMs expose predictive structure most clearly, auxiliary WAMs compress future information into a more VLA-like policy, and joint WAMs encode richer but more entangled trajectory-level information. At the same time, explicit future modeling introduces a non-trivial deployment cost, especially for inference-time imagination. Our findings suggest that the key question for future WAM design is not simply whether a model predicts the future, but whether its future representation is actionable. Effective WAMs should preserve enough future-oriented structure to improve target-object selectivity, distractor stability, and replanning consistency, while remaining efficient enough for closed-loop robotic control. This perspective reframes WAM evaluation: future prediction should not be judged only by visual plausibility or task success, but by whether it produces measurable improvements in robot behavior and interpretable predictive representations.


\vspace{-0.3cm}

\section{Related Work}
\label{sec:related_work}
\vspace{-0.2cm}

\begin{figure}[h]
    \centering
    \includegraphics[width=\linewidth]{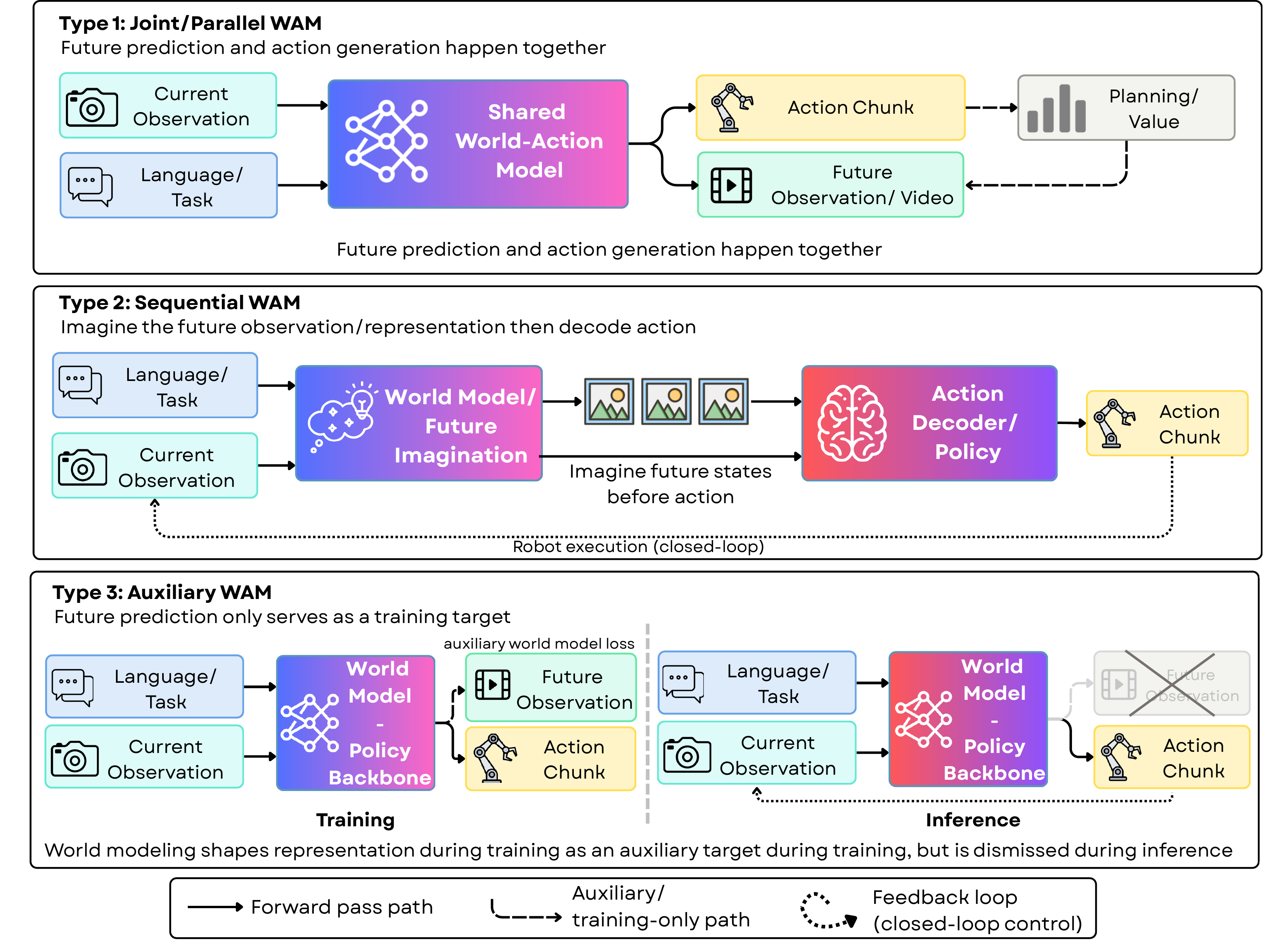}
  \caption{\small Overview of the three World-Action Model (WAM) paradigms.}
    \label{fig:wam_types}
\end{figure}

VLA policies such as RT-2 \cite{rt2}, $\pi_{0.5}$ \cite{pi05}, and Octo \cite{octo_2023} have become a central paradigm for generalist robot learning, mapping visual observations and language instructions to actions through scalable multimodal pretraining, language grounding, diffusion or flow-matching decoders, and cross-embodiment training \cite{chi2024diffusionpolicy,octo_2023,zheng2026xvla}. However, because most VLAs primarily predict actions without explicitly modeling how the scene should evolve, recent WAMs couple action generation with future imagination, latent dynamics, or world-modeling objectives, building on predictive world models \cite{worldmodel,gennie,dreamerv3} and recent visual world models such as V-JEPA2 \cite{assran2025vjepa2,vjepa21} or Cosmos \cite{cosmos}. Regarding Fig.~\ref{fig:wam_types}, we organize WAM-style methods into joint or parallel WAMs, which predict future observations and actions together, as in Cosmos Policy \cite{cosmospolicy} and DreamZero \cite{dreamzero}; sequential WAMs, which imagine future visual states before inferring actions, as in LingBot-VA \cite{lingbot-va2026} and Dream4Manip \cite{gu2026dream4manip}; and auxiliary or training-only WAMs, which use future prediction or alignment during training but remove explicit future rollout at inference, as in FastWAM \cite{yuan2026fastwam}, GigaWorld-Policy \cite{ye2026gigaworld}, VLA-JEPA \cite{vlajepa2026}, and FRAPPE \cite{zhao2026frappe}; we distinguish these from latent world-model planners such as LeWorldModel \cite{maes2026leworldmodel}, which select actions by predicting future latents under candidate actions with MPC or search. Since standard robot benchmarks (eg. LIBERO, DROID) \cite{rbm,liu2023libero,taomaniskill3,chen2025robotwin,droid,openx} often report only aggregate success, return, or completion time, which can hide differences in motion quality, replanning, object interaction, distractor disturbance, and representation use, recent work motivates more diagnostic evaluation of trajectories, action consistency, perturbation sensitivity, failure modes, nearest training examples, and feature structure \cite{chi2024diffusionpolicy,swann2026sparse,sentinel,agia2024unpacking}. Following this view, we compare WAMs and VLAs using both task outcomes and diagnostic measures of action dynamics, chunk-boundary consistency, object-state progress, manipulation-stage completion, wrong-object disturbance, latency, and learned feature representations.

\section{Evaluation Protocols}
\label{sec:method}

\subsection{Preliminary}
We study robot policies that map observations \(o_t \in \mathcal{O}\), robot or
environment states \(es_t \in \mathcal{S}\), and task instructions
\(g \in \mathcal{G}\) to executed actions \(a_t \in \mathcal{A}\). Many policies
predict action chunks
\(C_t=(a_t,\ldots,a_{t+H-1})\in\mathcal{A}^{H}\), where \(H\) is the chunk
horizon. A standard VLA policy directly maps the current observation and
instruction to actions, whereas a WAM incorporates future prediction during
training or inference. We consider three WAM families, illustrated in
Fig.~\ref{fig:wam_types}: \emph{joint/parallel WAMs} \cite{cosmospolicy}, which predict future
states and actions together; \emph{sequential WAMs} \cite{lingbot-va2026}, which first imagine future
observations or representations and then decode actions; and
\emph{auxiliary WAMs} \cite{vlajepa2026,yuan2026fastwam}, which use future prediction only as a training objective
and discard it at inference time. This taxonomy clarifies how future modeling
enters each policy.

\subsection{Behavioral Rollout Analysis}
\label{sec:behavior_metrics}

Analyzing the action space of robotic policies requires looking beyond rollout success and examining both the executed robot actions and their interactions with objects. Prior behavioral-diagnostic studies have emphasized this need, such as research in trajectory-planning \cite{agia2024unpacking}, runtime-monitoring~\cite{black2026realtime}, and action-consistency~\cite{GASPARETTO2007455,wang2026roboevalroboticmanipulationmeets}. Inheriting from this foundation, we consider three groups of diagnostics: standard outcome metrics (e.g., success rate); action-dynamics metrics, which characterize motion efficiency and stability; and object-interaction metrics, which measure whether the policy moves the target object while avoiding unnecessary distractor motion.

For action dynamics, we measure smoothness using the first-, second-, and third-order finite differences of the executed action sequence, corresponding to delta, acceleration, and jerk \cite{wang2026roboevalroboticmanipulationmeets}:
\[
\Delta a_t = a_t-a_{t-1}, \qquad \Delta^2 a_t = a_t-2a_{t-1}+a_{t-2}, \qquad \Delta^3 a_t = a_t-3a_{t-1}+3a_{t-2}-a_{t-3}.
\]
Intuitively, lower values of these metrics indicate smoother and more stable control. However, smoothness alone can be misleading on difficult benchmarks, since a failed policy may appear smooth simply because the robot barely moves (see Appendix~\ref{app:robotwin_audit}). To account for this case, we introduce low-motion failure, which measures failed episodes with very small average action change:
\[
M_{\mathrm{lowfail}}
=
 \frac{100}{N}\sum_{e=1}^{N}
\mathbf{1}\{y^{(e)}=0\}
\mathbf{1}\{\bar m^{(e)}<\tau_{\mathrm{motion}}\},
\qquad
where \ \ 
\bar m^{(e)}
=
\frac{1}{T_e-1}\sum_{t=2}^{T_e}\|\Delta a^{(e)}_t\|_2. 
\qquad 
\]
Besides, we measure whether a new policy chunk query causes an abrupt
change in the executed command. For valid action chunk query steps
\(\mathcal{Q}=\{q:s_q>1\}\), let
\(B_q=\left\|(a_{s_q}-a_{s_q-1})/\sigma_a\right\|_2\). We report the
episode-level mean boundary jump
\(M_{\mathrm{bdry}}=|\mathcal{Q}|^{-1}\sum_{q\in\mathcal{Q}}B_q\) and its
horizon-normalized variant
\(M_{\mathrm{bdry}/\sqrt{H}}=|\mathcal{Q}|^{-1}\sum_{q\in\mathcal{Q}}B_q/\sqrt{H}\). Also, we report auxiliary action-scale diagnostics, including normalized action magnitude \(\left\|a_t/\sigma_a|\right\|_2\) and sliding-window action standard deviation, which captures local action variability. Due to space constraints, full details on action-dynamics space metrics are deferred to Appendix~\ref{app:action_metrics}.


For object-interaction, we quantify robot-object interaction through object-level motion and
stability diagnostics. For example, we measure final progress, accumulated normalized progress, and the number of backward progress steps via goal positions \(p^{\mathrm{goal}}\) using
\(d_t=\|p^{\mathrm{tar}}_t-p^{\mathrm{goal}}_t\|_2\):
\[
M_{\mathrm{prog}}=d_0-d_T,\quad M_{\mathrm{progAUC}}=\frac{1}{T}\sum_{t=1}^{T}\frac{d_0-d_t}{d_0+\epsilon},\quad M_{\mathrm{rev}}=\sum_{t=1}^{T-1}\mathbf{1}\{d_{t+1}>d_t+\epsilon_{\mathrm{rev}}\}.
\]

To capture unintended interactions, we measure final distractor drift
and total distractor path length over distractor objects \(\mathcal{D}\) as
\[
M_{\mathrm{wrongmotion}}=\frac{1}{|\mathcal{D}|}\sum_{i\in\mathcal{D}}\|p^i_T-p^i_0\|_2,\quad M_{\mathrm{wrongpath}}=\frac{1}{|\mathcal{D}|}\sum_{i\in\mathcal{D}}\sum_{t=1}^{T-1}\|p^i_{t+1}-p^i_t\|_2.
\]

We also study other object-interaction metrics: net displacement, motion directness, and coarse manipulation stages. Due to space constraints, the full formal definitions of object-interaction metrics are fully described in Appendix~\ref{app:object_metrics}.

\subsection{Interpreting Feature-Space via Sparse Autoencoders}
\label{sec:sae_analysis}

Feature-space comparison between WAMs and VLAs is difficult due to polysemanticity~\cite{lan2025sparse} and architectural misalignment across models. Recent VLA interpretability work~\cite{buurmeijer2026observing, khan2025controlling} suggests that object-state and action information is often encoded in internal activations, but probe-based extraction alone does not provide a reusable feature basis for cross-model comparison. Hence, motivated by recent advances in mechanistic interpretability~\cite{lan2025sparse,gao2025scaling,swann2026sparse}, which demonstrate the effectiveness and high interpretability of Sparse Autoencoders (SAEs)~\cite{huben2024sparse} across both LLM and VLA studies, we adopt SAEs on comparable latent activation families, including \textbf{current visual latents}, \textbf{future imagined latents}, and \textbf{action latents}, for both VLAs and WAMs.
This gives each model its own sparse
feature basis while preserving a common set of feature-level diagnostics. The goal is to use SAEs as feature-level diagnostics to: 1) examine whether stronger object selectivity and target progress are associated with more future-oriented or temporally structured SAE statistics, and 2) test whether each policy family’s internal representations align with rollout-level behavioral differences.

\paragraph{SAEs.} Let \(x^{(e)}_t \in \mathbb{R}^d\) be the last or near-last hidden latent extracted from each modality-specific stream (vision, future, and action) of each model at timestep \(t\)
of episode \(e\). For each latent type of each model, we train a TopK SAE with an AuxK
auxiliary reconstruction loss. After centering and normalizing the activation to
\(\tilde{x}^{(e)}_t\), the SAE computes
\[
z^{(e)}_t =
\mathrm{ReLU}\!\left(\mathrm{TopK}(W_{\mathrm{enc}}\tilde{x}^{(e)}_t,k)\right),
\qquad
\hat{\tilde{x}}^{(e)}_t = W_{\mathrm{dec}}z^{(e)}_t ,
\]
where encoded latent \(z^{(e)}_{t,j}\) is the activation strength of SAE feature \(j\). Full normalization, optimization, and SAE-training details are given in
Appendix~\ref{app:sae_train}.

\paragraph{Metrics.} For each feature \(j\), we define \(f_j(x^{(e)}_t)=z^{(e)}_{t,j}\) and an active
state \(s^{(e)}_{t,j}=\mathbbm{1}[f_j(x^{(e)}_t)>\tau]\). Firstly, we summarize features'
activation pattern by inheriting 4 metrics proposed by Dr.VLA \cite{swann2026sparse}: episode coverage (\(c\)), onset count (\(\bar{o}\)), relative run length (\(\bar{\ell}\)), and
activation magnitude (\(\bar{a}\)). Further, to leverage our question concerning future modelling of WAM, we introduce 3 WAM-oriented futuristic metrics: \(\mathrm{FCS}\), \(\mathrm{HS}\), and \(\mathrm{AP}\). Specifically, Let \(d_j\) be the decoder direction of feature \(j\), and
let \(y^{(e)}_{t+\delta}\) denote a future latent state. The \textbf{Future Consistency
Score (\(\mathrm{FCS}\))} measures whether an active feature points toward future latent structure:
\[
\mathrm{FCS}_j =
\frac{1}{|\mathcal{A}_j|}
\sum_{(e,t)\in\mathcal{A}_j}
\max_{\delta\in[1,H_f]}
\operatorname{sim}_{\cos}\!\left(d_j,\tilde y^{(e)}_{t+\delta}\right),
\]
where \(\mathcal{A}_j=\{(e,t):s^{(e)}_{t,j}=1\}\). \textbf{Horizon Stability (\(\mathrm{HS}\))} measures
whether the feature remains coherent across future horizons, and\textbf{ Action
Predictiveness (\(\mathrm{AP}\))} measures how well its activation explains future action segments:
\[
\mathrm{HS}_j=\frac{1}{H}\sum{h=1}^{H}
\frac{\sum_e\sum_{t=1}^{T^{(e)}-h}s_{t,j}^{(e)}s_{t+h,j}^{(e)}}
{\sum_e\sum_{t=1}^{T^{(e)}-h}s_{t,j}^{(e)}+\epsilon},
\qquad
\mathrm{AP}_j =
R^2\!\left(
a^{(e)}_{t+1:t+H_a}
\leftarrow
f_j(x^{(e)}_t)
\right).
\]
Together, these metrics form a feature descriptor used to assign probabilistic
feature labels: \textit{predictive-general}, \textit{reactive-general}, or \textit{memorized}.
The full formal definitions of 7 metrics (\(c,\bar{o},\bar{\ell},\bar{a},\mathrm{FCS},\mathrm{HS},\mathrm{AP}\)) are provided in
Appendix~\ref{app:feature_metrics}.

\paragraph{Feature labelling \& classification.}
For these feature labels, \textbf{memorized} features capture meaningful but highly specific patterns, such as particular episodes, objects, or short trajectory fragments. \textbf{Reactive-general} features describe the current observation, robot state, or execution phase, such as grasping, lifting, or placing, and are shared across many episodes. \textbf{Predictive-general} features instead provide evidence about future states or actions through future-latent consistency or early activation before a later phase, or correlation with future action chunks across many episodes, especially in WAM future streams. Then, given
\(m_j=[c_j,\bar{o}_j,\bar{\ell}_{r,j},\bar{a}_j,\mathrm{FCS}_j,\mathrm{HS}_j,\mathrm{AP}_j]\),
we train a class-balanced multinomial logistic classifier on the manually audited features sample set as
$
P(y_j=k\mid m_j)=\sigma(w_k^\top m_j+b_k),
$
which is used for classifying all non-dead SAE features across all models for each dataset. Full labelling and auditing details are
provided in Appendix~\ref{app:feature_labeling}.


\section{Experiment}
\label{sec:result}

\subsection{Experimental Setup}

We evaluate 7 representative VLA and WAM robot policies on action-space behavior, feature-space
structure, and computational cost on a single RTX 6000 Ada GPU. The evaluated
VLAs include $\pi_0$ \cite{pi0}, $\pi_{0.5}$ \cite{pi05}, and X-VLA
\cite{zheng2026xvla}; since $\pi_0$ was not originally evaluated in simulation,
we use the LeRobot-finetuned checkpoint \cite{cadene2026lerobot} instead. The WAMs cover the three design families:
Cosmos Policy \cite{cosmospolicy} as a joint WAM, LingBot-VA
\cite{lingbot-va2026} as a sequential WAM, and FastWAM
\cite{yuan2026fastwam} plus VLA-JEPA \cite{vlajepa2026} as auxiliary WAMs. We
evaluate on LIBERO \cite{liu2023libero} and RoboTwin2.0
\cite{chen2025robotwin}. On LIBERO, all seven policies are evaluated on the
spatial, object, goal, and long suites with 50 trials per task, yielding 2,000
episodes; LingBot-VA is evaluated only on LIBERO-10 because this is the released
checkpoint. On RoboTwin2.0, we evaluate the public-checkpoint subset
(LingBot-VA, FastWAM, \(\pi_{0.5}\)) on 50 tasks, two setups per task,
and 10 rollouts per setup, also yielding 2,000 episodes.

\begin{table*}[h]
\centering
\scriptsize
\setlength{\tabcolsep}{2.6pt}
\renewcommand{\arraystretch}{0.95}
\caption{\small \textbf{LIBERO task success and motion regularity.}
Lower delta, acceleration, jerk, boundary jump, boundary/\(\sqrt{H}\), window
standard deviation, and low-motion failure indicate better executed control.
Row colors denote WAM subtype respectively: \colorbox{jointwam}{joint WAM}, \colorbox{seqwam}{sequential WAM}, and
\colorbox{auxwam}{auxiliary WAM}.}
\label{tab:libero_motion}
\resizebox{\textwidth}{!}{
\begin{tabular}{clrrrrrrrrrr}
\toprule
&
Model 
& \makecell{Success\\\(\uparrow\)} 
& \makecell{Ep. len.\\\(\downarrow\)} 
& \makecell{Delta\\L2 \(\downarrow\)}
& \makecell{Accel\\L2 \(\downarrow\)}
& \makecell{Jerk\\L2 \(\downarrow\)}
& \makecell{Boundary\\jump \(\downarrow\)}
& \makecell{Boundary\\/\(\sqrt{H}\) \(\downarrow\)}
& \makecell{Window\\std. \(\downarrow\)} 
& \makecell{Norm. action\\L2}
& \makecell{Low-motion\\fail. \((\%)\downarrow\)} \\
\midrule
\rowcolor{jointwam}
& Cosmos     & \textbf{98.0} & 153 & 0.104 & 0.104 & 0.185 & 0.387 & 0.097 & 0.219 & 1.489 & 0.5 \\
\rowcolor{seqwam}
& Lingbot-VA & \textbf{98.0} & 276 & \textbf{0.079} & \textbf{0.093} & \textbf{0.173} & \textbf{0.153} & 0.038 & \textbf{0.170} & \textbf{1.438} & 0.8 \\
\rowcolor{auxwam}
& FastWAM    & 95.3 & 161 & 0.096 & 0.098 & 0.176 & 0.298 & 0.053 & 0.212 & 1.515 & 0.9 \\
\rowcolor{auxwam}
\multirow{-4}{*}{\rotatebox[origin=c]{90}{\textsc{WAM}}}
& VLA-JEPA   & 97.1 & \textbf{151} & 0.110 & 0.117 & 0.208 & 0.156 & 0.059 & 0.226 & 1.507 & \textbf{0.2} \\
\midrule
\multirow{3}{*}{\rotatebox[origin=c]{90}{\textsc{VLA}}}
& \(\pi_0\)     & 66.9 & 220 & 0.141 & 0.199 & 0.367 & 0.666 & 0.094 & 0.257 & 1.475 & 9.3 \\
& \(\pi_{0.5}\) & 96.6 & 160 & 0.113 & 0.124 & 0.220 & 0.298 & 0.094 & 0.227 & 1.493 & 1.3 \\
& X-VLA         & 95.8 & 178 & 0.136 & 0.245 & 0.485 & 0.161 & \textbf{0.029} & 0.368 & 1.910 & 0.9 \\
\bottomrule
\end{tabular}}
\end{table*}

\begin{table*}[h]
\centering
\scriptsize
\setlength{\tabcolsep}{3pt}
\renewcommand{\arraystretch}{0.95}
\caption{\small \textbf{LIBERO object interaction and distractor stability.}
Object metrics are computed on episodes with decoded target-object states.
Wrong-object metrics report mean distractor displacement or path length.}
\label{tab:libero_object}
\resizebox{\textwidth}{!}{
\begin{tabular}{clrrrrrrrrrrr}
\toprule
&
Model
& \makecell{Prog.\\\(\uparrow\)}
& \makecell{Prog.\\AUC \(\uparrow\)}
& \makecell{Rev.\\\(\downarrow\)}
& \makecell{Target\\path \(\downarrow\)}
& \makecell{Direct.\\\(\uparrow\)}
& \makecell{Wrong\\motion \(\downarrow\)}
& \makecell{Wrong\\path \(\downarrow\)}
& \makecell{Grasp\\(\%) \(\uparrow\)}
& \makecell{Lift\\(\%) \(\uparrow\)}
& \makecell{Transport\\(\%) \(\uparrow\)}
& \makecell{Place\\(\%) \(\uparrow\)} \\
\midrule
\rowcolor{jointwam}
& Cosmos
& 0.166 & 0.084 & 23.8 & 0.314 & \textbf{0.744}
& \textbf{0.011} & \textbf{0.017} & 52.5 & 50.2 & 63.2 & 41.8 \\

\rowcolor{seqwam}
& Lingbot-VA
& \textbf{0.199} & \textbf{0.240} & \textbf{2.4} & 0.413 & 0.668
& 0.157 & 0.174 & 69.4 & \textbf{66.8} & 50.7 & \textbf{50.0} \\

\rowcolor{auxwam}
& FastWAM
& 0.127 & 0.057 & 37.2 & 0.482 & 0.456
& 0.212 & 0.673 & \textbf{68.3} & 54.2 & 62.7 & 40.5 \\

\rowcolor{auxwam}
\multirow{-4}{*}{\rotatebox[origin=c]{90}{\textsc{WAM}}}
& VLA-JEPA
& 0.162 & 0.062 & 22.3 & \textbf{0.304} & 0.407
& 0.030 & 0.036 & 52.0 & 49.6 & \textbf{64.3} & 40.4 \\
\midrule
\multirow{3}{*}{\rotatebox[origin=c]{90}{\textsc{VLA}}}
& \(\pi_0\)
& 0.074 & 0.029 & 74.0 & 0.446 & 0.415
& 0.225 & 0.814 & 61.6 & 42.5 & 48.9 & 30.0 \\

& \(\pi_{0.5}\)
& 0.128 & 0.056 & 53.2 & 0.497 & 0.647
& 0.211 & 0.687 & 67.6 & 54.1 & 64.2 & 42.1 \\

& X-VLA
& 0.119 & 0.047 & 39.3 & 0.481 & 0.442
& 0.220 & 0.736 & 68.0 & 53.6 & 61.7 & 39.4 \\
\bottomrule
\end{tabular}}
\end{table*}

\subsection{Action-Space Results}
\label{sec:action_space_results}


\paragraph{LIBERO.}
Table~\ref{tab:libero_motion} shows that task success alone does not tell the
whole story, especially on a high-success benchmark like LIBERO, the motion diagnostics
reveal a clearer separation between the two families. WAM policies
generally produce smoother executed control, with lower acceleration, jerk,
boundary discontinuity, and low-motion failure rates. For further explainability, Table~\ref{tab:libero_object} helps clarify this difference at the object level.
The object metrics show that WAMs generally achieve stronger target progress,
more direct object movement, and lower wrong-object disturbance, indicating
better spatial grounding and target-object selectivity during manipulation.
Auxiliary WAMs are especially informative in this case: since these inference-time policies remain
closer to a direct VLA, their object-level behavior often lies between explicit
inference-time WAMs and VLAs. This supports the importance of imagination during
inference of WAMs. Although auxiliary WAMs designs try to omit the imagination phase to
reduce computational cost, our results suggest that doing so can also reduce the
spatial-awareness benefits provided by future object prediction.

\begin{wraptable}[10]{r}{0.5\textwidth}
\vspace{-0.3cm}
\centering
\scriptsize
\setlength{\tabcolsep}{3pt}
\renewcommand{\arraystretch}{0.92}
\caption{\small Runtime and deployment cost on LIBERO.}
\label{tab:runtime}
\resizebox{\linewidth}{!}{
\begin{tabular}{lrrrrr}
\toprule
Model 
& \makecell{p50\\(ms) \(\downarrow\)} 
& \makecell{p95\\(ms) \(\downarrow\)}
& \makecell{Chunks/s\\\(\uparrow\)}
& \makecell{Eff. Hz\\\(\uparrow\)}
& \makecell{GPU\\(MB)} \\
\midrule
Cosmos     &  956 &  977 & 1.049 & 16.780 &  9,762 \\
FastWAM    & 1418 & 1529 & 0.681 & 6.810 & 27,891 \\
LingBot-VA & 4701 & 4911 & 0.216 &  3.460 & 24,130 \\
VLA-JEPA   & 1588 & 1738 & 0.569 &  3.982 &  5,462 \\
\midrule
\(\pi_0\)     & 194 & 228 & 4.348 & \textbf{217.381} & 8,631 \\
\(\pi_{0.5}\) & \textbf{100} & \textbf{103} & \textbf{7.096} &  70.963 & 9,124 \\
X-VLA         & 329 & 360 & 6.895 & 29.850 & \textbf{7,089} \\
\bottomrule
\end{tabular}}
\vspace{-0.6cm}
\end{wraptable}

\paragraph{Runtime trade-off.}
Table~\ref{tab:runtime} highlights a well-known deployment trade-off: WAMs are
often substantially more expensive at inference than direct VLA policies.
This is expected because WAMs require additional computation to imagine future
states or rollouts before selecting actions, with the cost especially visible for
sequential WAMs (Lingbot-VA). Thus, the behavioral and object-space advantages of WAM-like
policies do not come for free. A key open problem is how to reduce this inference
cost while preserving the benefits of future imagination.

\par\noindent

\vspace{-0.2cm}

\begin{table*}[h]
\centering
\scriptsize
\setlength{\tabcolsep}{3pt}
\renewcommand{\arraystretch}{0.95}
\caption{\small \textbf{RoboTwin2.0 task success and motion regularity.}
RoboTwin2.0 uses dual-arm qpos targets, so these values are not directly comparable
with LIBERO delta-action metrics. Lower delta, acceleration, jerk, boundary jump,
boundary/\(\sqrt{H}\), window standard deviation, and low-motion failure indicate
better executed control.}
\label{tab:robotwin_main}
\resizebox{\textwidth}{!}{
\begin{tabular}{lrrrrrrrrrr}
\toprule
Model 
& \makecell{Success\\\(\uparrow\)} 
& \makecell{Ep. len.\\\(\downarrow\)} 
& \makecell{Delta\\L2 \(\downarrow\)}
& \makecell{Accel\\L2 \(\downarrow\)}
& \makecell{Jerk\\L2 \(\downarrow\)}
& \makecell{Boundary\\jump \(\downarrow\)}
& \makecell{Boundary\\/\(\sqrt{H}\) \(\downarrow\)}
& \makecell{Window\\std. \(\downarrow\)} 
& \makecell{Norm. action\\L2}
& \makecell{Low-motion\\fail. \((\%)\downarrow\)} \\
\midrule
LingBot-VA    & \textbf{81.6} & \textbf{318} & \textbf{0.030} & 0.020 & 0.036 & 0.065 & 0.012 & \textbf{0.067} & 2.310 &  \textbf{6.1} \\
FastWAM       & 67.3 & 367 & 0.052 & 0.025 & 0.043 & 0.038 & 0.007 & 0.110 & 2.066 & 32.7 \\
\midrule
\(\pi_{0.5}\) & 44.3 & 410 & 0.044 & \textbf{0.015} & \textbf{0.022} & \textbf{0.011} & \textbf{0.002} & 0.100 & 2.043 & 53.0 \\
\bottomrule
\end{tabular}}
\vspace{-0.3cm}
\end{table*}

\paragraph{RoboTwin2.0.} Table~\ref{tab:robotwin_main} shows a different regime from LIBERO: overall
success is substantially lower, and the motion metrics appear at first glance to
follow the opposite trend. However, this should be interpreted together with the
low-motion failure rate. The very high low-motion failure of \(\pi_{0.5}\)
indicates that many of its failed rollouts involve little effective robot
operation, which can make the trajectory appear artificially smooth (see Appendix~\ref{app:robotwin_audit}, Table~\ref{tab:robotwin_success_cond}). Thus, lower
acceleration, jerk, or boundary jump in this setting does not necessarily imply
better control. Instead, RoboTwin result reinforces three points: 1) WAMs maintain reasonable smoothness while actively manipulating objects, 2) WAMs therefore still benefit
from imagination under harder tasks,  and 3) motion metrics are most informative when
success rates are high or nearly comparable across models, such as those results in Table~\ref{tab:libero_motion}.


\begin{table*}
\centering
\scriptsize
\setlength{\tabcolsep}{3.5pt}
\renewcommand{\arraystretch}{0.95}
\caption{\small \textbf{Per-model SAE feature classification by representation space.}
\#Active Feat. denotes the number of active non-dead SAE features. Dead percentage is
computed over all learned SAE features. Percentages for memorized, reactive, and
predictive features are computed over non-dead features with joint weak labels.
\bestcurrent{Red}, \bestfuture{orange}, and \bestaction{blue} indicate the highest value among current, future, and action features, respectively.}
\label{tab:model_space_feature_summary}
\resizebox{0.65\linewidth}{!}{
\begin{tabular}{cllrrrrr}
\toprule
&
Model
& Space
& \makecell{\#Active \\ Feat.}
& \makecell{Dead\\(\%)}
& \makecell{\textbf{Mem.}\\(\%)}
& \makecell{\textbf{React.}\\(\%)}
& \makecell{\textbf{Pred.}\\(\%)} \\
\midrule

\rowcolor{jointwam}
&        & Current & 12,122 & 1.4 & 83.7 & 13.6 & 2.7 \\
\rowcolor{jointwam}
& Cosmos & Future  & 11,811 & 3.9 & \bestfuture{80.0} & 14.1 & 5.9 \\
\rowcolor{jointwam}
&        & Action  & 4,093  & 0.1 & \bestaction{84.9} & 13.1 & 2.0 \\

\noalign{\vskip 2pt}

\rowcolor{seqwam}
&            & Current & 2,170 & 64.7  & 10.3 & \bestcurrent{72.7} & \bestcurrent{17.0} \\
\rowcolor{seqwam}
& LingBot-VA & Future  & 5,997 & 2.4 & 5.0 & 41.8 & \bestfuture{53.2} \\
\rowcolor{seqwam}
&            & Action  & 376   & 81.6 & 30.1 & \bestaction{65.6} & 4.3 \\

\noalign{\vskip 2pt}

\rowcolor{auxwam}
&         & Current & 583   & 85.8 & 25.2 & \bestfuture{70.8} & 4.0 \\
\rowcolor{auxwam}
& FastWAM & Future  & 246   & 94.0 & 2.0  & 90.7 & 7.3 \\
\rowcolor{auxwam}
&         & Action  & 2,707 & 11.9 & 75.9 & 22.1 & 2.0 \\

\noalign{\vskip 2pt}

\rowcolor{auxwam}
&          & Current & 3,957 & 3.4  & 76.0 & 10.8 & 13.2 \\
\rowcolor{auxwam}
& VLA-JEPA & Future  & 2,537 & 38.1 & 59.8 & 23.3 & 16.9 \\
\rowcolor{auxwam}
\multirow{-12}{*}{\rotatebox[origin=c]{90}{\textsc{WAM}}}
&          & Action  & 92    & 91.0 & 2.2  & 38.0 & \bestaction{59.8} \\

\midrule

& $\pi_0$     & Current & 1,310 & 68.0 & 67.4 & 31.8 & 0.8 \\
&             & Action  & 1,021 & 0.3  & 52.3 & 47.2 & 0.5 \\

\noalign{\vskip 2pt}

& $\pi_{0.5}$ & Current & 5,108 & 37.6 & 64.1 & 34.7 & 1.2 \\
&             & Action  & 242   & 88.2 & 44.4 & 54.8 & 0.8 \\

\noalign{\vskip 2pt}

& X-VLA       & Current & 2,028 & 1.0  & \bestcurrent{99.4} & 0.3  & 0.3 \\
\multirow{-6}{*}{\rotatebox[origin=c]{90}{\textsc{VLA}}}
&             & Action  & 529   & 74.2 & 36.5 & 62.6 & 0.9 \\

\bottomrule
\end{tabular}}
\vspace{-0.5em}
\end{table*}

\subsection{Feature-Space Result}

Table~\ref{tab:model_space_feature_summary} shows that future-oriented SAE
features are strongly architecture-dependent. Direct VLA baselines have extremely low predictive features rate, while WAMs expose a non-zero predictive structure,
with LingBot-VA showing the clearest future-oriented representation due to its
sequential imagine-then-act design. Auxiliary WAMs are more mixed: VLA-JEPA
retains a noticeably larger predictive share than FastWAM, which is consistent
with its cleaner object interaction in Table~\ref{tab:libero_object}. More
broadly, comparing the SAE statistics with the object metrics (eg. Lingbot-VA, VLA-JEPA \& FastWAM) suggests that
models with stronger future-visualization capacity, whether used during training
or inference, tend to have better spatial grounding and target-object
selectivity. LingBot-VA illustrates this most clearly: even when its absolute
number of manually audited predictive features is not larger than VLA-JEPA in
Table~\ref{tab:manual_audit_counts}, predictive features form a dominant
proportion of its active representation, matching both its model design and its
rollout behavior. These results
support the view that future-oriented representations help explain the
object-level advantages of WAMs, we further enhance this view in Appendix~\ref{app:sae_behavior_probe}.

\vspace{-0.3cm}

\begin{figure}[h]
    \centering
    \includegraphics[width=\linewidth]{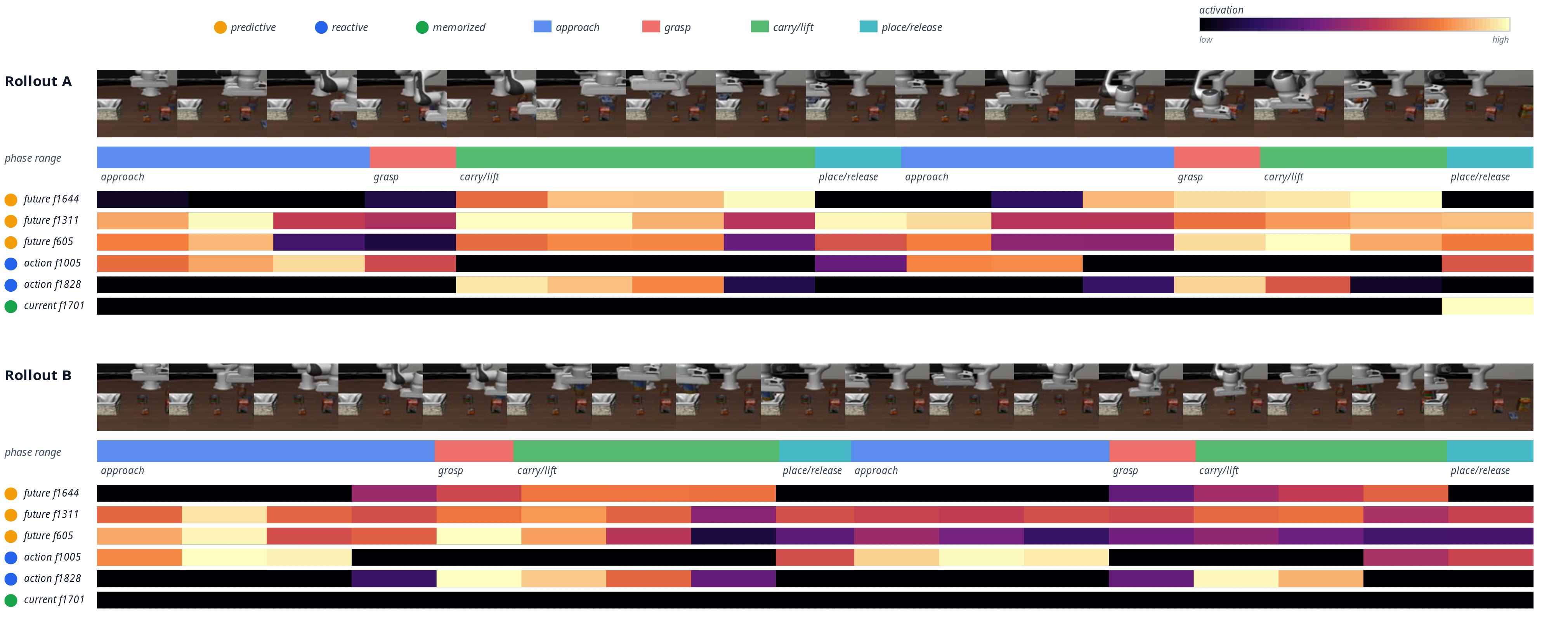}
    \caption{\small \textbf{Phase-aligned sparse feature activations.}
    2 Cosmos LIBERO-10 rollouts are shown with task phases and selected SAE
    activations. General features show phase-specific patterns, e.g.,
    \(f_{1644}\) during carry/lift and \(f_{1828}\) during grasp, while the
    memorized feature \(f_{1701}\) appears only in one rollout near
    place/release.}
    \label{fig:cosmos_libero_phase_features}
    \vspace{-0.3cm}
\end{figure}

\paragraph{Phase alignment.}
To further validate SAE features, we visualized the temporal phase alignment to a set of sampled features as shown in Fig.~\ref{fig:cosmos_libero_phase_features}. Predictive features tend to
activate before or across phase transitions, suggesting sensitivity to upcoming
task stages, while reactive features concentrate around execution-heavy phases
such as grasp and carry. Memorized features are sparse and episode-specific.
This visualization supports the quantitative SAE analysis, and again confirms the high interpretability of survived SAE features, which align with prior works \cite{lan2025sparse,swann2026sparse}.

\subsection{Discussion}

Our results show that comparing WAMs and VLAs only by success rate misses an
important part of the story. The action-space and object-space diagnostics reveal
that WAMs tend to produce smoother and more object-oriented behavior, with better
target-object selectivity and fewer unintended interactions. The SAE analysis
provides a complementary feature-space view: WAMs expose more future-oriented
representations than direct VLAs, and these representations are associated with
stronger object-level behavior. While this evidence is still associative rather
than being causal, it partially explains why future prediction help WAMs
achieve better spatial grounding and cleaner object interaction.

Meanwhile, the runtime results show that these benefits come with a clear
deployment cost, and therefore, balancing cost and future imagination is non-trivial. A key direction for future work is therefore to make WAMs more
efficient without removing the property that makes them useful: explicit future
prediction. Sequential WAMs expose future information clearly but are expensive, while joint
WAMs such as Cosmos reduce the explicit separation between imagination and action
and can make the representation more entangled. Auxiliary designs are efficient,
but they risk compressing or weakening the future-modeling signal at inference
time. This requires a better balance between sequential and joint designs.
Future architectures should therefore preserve lightweight inference-time
imagination, rather than treating future prediction only as a training auxiliary
loss. For example, latent video world models such as V-JEPA 2-style architectures \cite{vjepa21,assran2025vjepa2selfsupervisedvideo} suggest a promising path: if future latent prediction can be used directly during
policy inference, rather than only as a vision encoder or auxiliary target, WAMs
may retain their spatial-awareness benefits while reducing runtime cost.

\section{Conclusion \& Limitation}
\label{sec:conclusion}
In conclusion, our study shows that WAMs and VLAs differ in ways that are not captured by task success alone. By combining rollout-level behavioral diagnostics with SAE-based feature analysis, we find that WAMs often produce stronger target-object progress, lower distractor disturbance, and more future-oriented internal representations, suggesting that explicit future modeling can be behaviorally useful for manipulation. However, our evidence remains associative rather than causal: SAE feature labels are weak probabilistic annotations, object-state metrics depend on available decoded traces, and the evaluated models are limited to simulation benchmarks and public checkpoints. In addition, WAMs introduce a clear runtime cost, highlighting the need for future designs that preserve actionable future prediction while remaining efficient for closed-loop robot control.

\clearpage


\bibliography{example}  

\newpage

\appendix

\section{High-Level Behavioral Metrics}
\label{app:high_level_metrics}

This appendix defines the rollout metrics used in our behavioral analysis. Each
rollout is converted into a canonical episode record containing the executed
action stream \(a_{1:T}\), optional action chunks
\(C \in \mathbb{R}^{Q \times H \times A}\), policy-query timesteps
\(\{s_q\}_{q=1}^{Q}\), robot and end-effector states, reward or success traces,
object-state traces when available, and runtime measurements. This common record
allows metrics to be computed across policies with different action adapters,
chunk horizons, replanning schedules, and logging formats. Metrics are computed
only when the required streams are available; for example, object metrics are
computed only for episodes with decoded object trajectories.

\subsection{Action Metrics}
\label{app:action_metrics}

\paragraph{Outcome metrics.}
We report standard task-level outcomes including success rate, episode length,
final step, completion ratio, return, final reward, maximum reward, and
first-success timestep when reward traces are available. These metrics indicate
whether a policy solves the task, but they do not explain how the policy moves,
how it replans, or why it fails.

\paragraph{Executed action magnitude.}
Given executed actions \(a_t \in \mathbb{R}^{A}\), we compute the raw action
magnitude using \(\|a_t\|_2\). When comparing policies with different action
scales or adapters, we use a global action scale \(\sigma_a\) and report
\[
M_{\mathrm{act}}
=
\frac{1}{T}\sum_{t=1}^{T}
\left\|\frac{a_t}{\sigma_a}\right\|_2 .
\]
When available, we also compute separate translation, rotation, and gripper
components.

\paragraph{Action smoothness.}
We measure local action smoothness using finite differences of the executed
action stream:
\[
\begin{aligned}
\Delta a_t &= a_t-a_{t-1}, &
\Delta^2 a_t &= a_t-2a_{t-1}+a_{t-2}, &
\Delta^3 a_t &= a_t-3a_{t-1}+3a_{t-2}-a_{t-3}.
\end{aligned}
\]
These correspond to executed action change, acceleration, and jerk. For each
episode, we summarize them by mean L2 magnitude:
\[
\begin{aligned}
M_{\Delta}
&= \frac{1}{T-1}\sum_{t=2}^{T}\|\Delta a_t\|_2, &
M_{\Delta^2}
&= \frac{1}{T-2}\sum_{t=3}^{T}\|\Delta^2 a_t\|_2, &
M_{\Delta^3}
&= \frac{1}{T-3}\sum_{t=4}^{T}\|\Delta^3 a_t\|_2 .
\end{aligned}
\]
Lower values indicate smaller step-to-step changes, fewer abrupt control
changes, and less high-frequency oscillation. We also report median, p95, and
maximum values when appropriate.

\paragraph{Sliding-window action variation.}
To capture short-horizon instability, we compute a sliding-window action
standard deviation. For window size \(w\), let
\(\tilde a_t=a_t/\sigma_a\) and
\(\mathcal{W}_t=\{\tilde a_{t-w+1},\ldots,\tilde a_t\}\). The window variation
score is
\[
M_{\mathrm{win}}
=
\frac{1}{T-w+1}\sum_{t=w}^{T}
\left\|\mathrm{Std}(\mathcal{W}_t)\right\|_2 .
\]
This metric captures local action variability that may be hidden by
episode-level averages.

\paragraph{Boundary discontinuity.}
For chunked policies, we measure whether a new policy query causes an abrupt
change in the executed command. At query step \(s_q\), define
\(B_q=\left\|(a_{s_q}-a_{s_q-1})/\sigma_a\right\|_2\). The episode-level
boundary scores are
\[
\begin{aligned}
M_{\mathrm{bdry}}
&=
\frac{1}{|\mathcal{Q}|}\sum_{q\in\mathcal{Q}}B_q, &
M_{\mathrm{bdry}/\sqrt{H}}
&=
\frac{1}{|\mathcal{Q}|}\sum_{q\in\mathcal{Q}}\frac{B_q}{\sqrt{H}},
\end{aligned}
\]
where \(\mathcal{Q}\) contains valid query steps with \(s_q>1\). When needed, we
also compute the boundary-to-non-boundary ratio
\[
R_{\mathrm{bdry}}
=
\frac{
|\mathcal{Q}|^{-1}\sum_{q\in\mathcal{Q}}B_q
}{
|\mathcal{T}_{\mathrm{non}}|^{-1}
\sum_{t\in\mathcal{T}_{\mathrm{non}}}
\left\|(a_t-a_{t-1})/\sigma_a\right\|_2+\epsilon
},
\]
where \(\mathcal{T}_{\mathrm{non}}\) denotes non-boundary timesteps.

\paragraph{Low-motion failure.}
Smoothness metrics can be misleading when a failed policy barely moves. For
episode \(e\), define its mean action displacement as
\[
\bar m^{(e)}
=
\frac{1}{T_e-1}\sum_{t=2}^{T_e}\|\Delta a^{(e)}_t\|_2 .
\]
The low-motion failure rate is
\[
M_{\mathrm{lowfail}}
=
100\cdot \frac{1}{N}\sum_{e=1}^{N}
\mathbf{1}\{y^{(e)}=0\}
\mathbf{1}\{\bar m^{(e)}<\tau_{\mathrm{motion}}\}.
\]
This counts failed episodes in which the policy produces very little motion.

\paragraph{Runtime and deployment cost.}
We report policy inference latency, client round-trip latency, server inference
latency, throughput in chunks per second, effective control frequency, real-time
ratio, and GPU memory usage. Latency is summarized using mean, p50, and p95.
These metrics capture whether a policy is practical to deploy, rather than only
whether it achieves high success.

\subsection{Object Metrics}
\label{app:object_metrics}

Object metrics are computed when decoded object trajectories are available. Let
\(p^{\mathrm{tar}}_t\) denote the target-object position at timestep \(t\), and
let \(p^{\mathrm{goal}}_t\) denote the goal or receptacle position. We define
the target-goal distance as
\[
d_t=\left\|p^{\mathrm{tar}}_t-p^{\mathrm{goal}}_t\right\|_2 .
\]
For RoboTwin2.0, the same definitions are applied to mapped reference objects;
therefore, progress metrics should be interpreted as benchmark-local
reference-object diagnostics.

\paragraph{Target/reference progress.}
Progress measures the final improvement in object-to-goal distance,
\(M_{\mathrm{prog}}=d_0-d_T\). Higher values indicate that the object ends
closer to the goal.

\paragraph{Progress AUC.}
Progress AUC measures whether the object stays closer to the goal throughout the
trajectory, rather than only at the final timestep:
\[
M_{\mathrm{progAUC}}
=
\frac{1}{T}\sum_{t=1}^{T}\frac{d_0-d_t}{d_0+\epsilon}.
\]
Higher values indicate steadier object-level progress.

\paragraph{Progress reversals.}
Progress reversals count how often the target object moves farther from the goal:
\[
M_{\mathrm{rev}}
=
\sum_{t=1}^{T-1}
\mathbf{1}\{d_{t+1}>d_t+\epsilon_{\mathrm{rev}}\}.
\]
Lower values indicate more monotonic manipulation.

\paragraph{Target-object path and directness.}
We measure target-object efficiency using path length, net displacement, and
directness:
\[
\begin{aligned}
L_{\mathrm{tar}}
&=
\sum_{t=1}^{T-1}
\left\|p^{\mathrm{tar}}_{t+1}-p^{\mathrm{tar}}_t\right\|_2, &
D_{\mathrm{tar}}
&=
\left\|p^{\mathrm{tar}}_T-p^{\mathrm{tar}}_0\right\|_2, &
\mathrm{Dir}_{\mathrm{tar}}
&=
\frac{D_{\mathrm{tar}}}{L_{\mathrm{tar}}+\epsilon}.
\end{aligned}
\]
Lower path length and higher directness indicate more efficient object movement.

\paragraph{Wrong-object motion.}
Let \(\mathcal{D}\) denote the set of distractor or non-target objects, and let
\(p^i_t\) be the position of distractor \(i\). We measure unintended distractor
movement using final displacement and full trajectory path length:
\[
\begin{aligned}
M_{\mathrm{wrongmotion}}
&=
\frac{1}{|\mathcal{D}|}
\sum_{i\in\mathcal{D}}
\left\|p^i_T-p^i_0\right\|_2, &
M_{\mathrm{wrongpath}}
&=
\frac{1}{|\mathcal{D}|}
\sum_{i\in\mathcal{D}}\sum_{t=1}^{T-1}
\left\|p^i_{t+1}-p^i_t\right\|_2 .
\end{aligned}
\]
Lower values indicate fewer unintended interactions with non-target objects.
The path metric also captures distractors that move during execution even if
they later return close to their initial positions.

\paragraph{Manipulation-stage completion.}
We report grasp, lift, transport, and place completion rates:
\[
M_{\mathrm{stage}}
=
100\cdot \frac{1}{N}\sum_{e=1}^{N}
\mathbf{1}\{\mathcal{C}_{\mathrm{stage}}^{(e)}\},
\]
where \(\mathcal{C}_{\mathrm{stage}}^{(e)}\) is the stage-specific completion
condition for episode \(e\). Grasp indicates that the target object is held, lift
indicates sufficient vertical displacement, transport indicates sufficient
movement toward the goal, and place indicates final placement within a goal
threshold.

\section{Details on Feature-Space Analysis}
\label{app:sae_detail}

\subsection{Details on Sparse Autoencoder}
\label{app:sae_train}

For each model and latent family, we collect hidden activations
\(x^{(e)}_t \in \mathbb{R}^d\), where \(e\) indexes episodes and \(t\) indexes
timesteps. Each activation is centered by a learned pre-bias \(b_{\rm pre}\),
then normalized by subtracting the per-sample mean and scaling by its
\(\ell_2\) norm:
\[
\begin{aligned}
x^{(e)}_{c,t}
&=x^{(e)}_t-b_{\rm pre}, &
\tilde{x}^{(e)}_t
&=
\frac{x^{(e)}_{c,t}-\mu(x^{(e)}_{c,t})}
{\|x^{(e)}_{c,t}-\mu(x^{(e)}_{c,t})\|_2+\epsilon}.
\end{aligned}
\]
The encoder maps this normalized activation into a sparse feature vector, and
the decoder reconstructs the normalized activation:
\[
\begin{aligned}
h^{(e)}_t
&=W_{\rm enc}\tilde{x}^{(e)}_t, &
z^{(e)}_t
&=\mathrm{ReLU}\!\left(\mathrm{TopK}(h^{(e)}_t,k)\right), &
\hat{\tilde{x}}^{(e)}_t
&=W_{\rm dec}z^{(e)}_t .
\end{aligned}
\]
Here \(\mathrm{TopK}(\cdot,k)\) keeps only the \(k\) largest coordinates and
sets the remaining coordinates to zero. Decoder columns are unit-normalized so
that each SAE latent corresponds to a feature direction and \(z^{(e)}_{t,j}\)
gives the activation strength of feature \(j\).

The SAE objective combines normalized reconstruction loss with an auxiliary
residual reconstruction branch over inactive or underused latents:
\[
\begin{aligned}
\mathcal{L}_{\rm rec}
&=
\|\tilde{x}^{(e)}_t-\hat{\tilde{x}}^{(e)}_t\|_2^2, &
r^{(e)}_t
&=
\tilde{x}^{(e)}_t-\hat{\tilde{x}}^{(e)}_t,\\
\mathcal{L}_{\rm aux}
&=
\|r^{(e)}_t-\hat r^{(e)}_{t,\rm aux}\|_2^2, &
\mathcal{L}_{\rm SAE}
&=
\mathcal{L}_{\rm rec}+\alpha\mathcal{L}_{\rm aux},
\quad \alpha=\frac{1}{32}.
\end{aligned}
\]
This objective encourages sparse feature activations while allowing unused
latents to specialize to remaining reconstruction error.

\subsection{Feature-Space Metrics}
\label{app:feature_metrics}

This appendix gives the full definition of the feature-space metrics used in
Section~\ref{sec:sae_analysis}. The first four metrics follow the
activation-pattern statistics used in Dr.VLA~\cite{swann2026sparse}, while the
remaining metrics are introduced to measure future-oriented structure in WAM
representations.

\paragraph{Notation.}
Let \(E=\{1,\ldots,N\}\) be the set of evaluation episodes, and let episode
\(e\) have length \(T^{(e)}\). For SAE feature \(j\), its activation and binary
activity indicator are
\[
f_j(x_t^{(e)})=z_{t,j}^{(e)}, \qquad
s_{t,j}^{(e)}=\mathbf{1}\!\left[f_j(x_t^{(e)})>\tau\right].
\]
The active-episode set for feature \(j\) is
\[
E_j^+
=
\left\{
e\in E:
\exists t\in\{1,\ldots,T^{(e)}\}
\text{ such that } s_{t,j}^{(e)}=1
\right\}.
\]
Unless otherwise stated, all averages below are computed only over active
episodes \(E_j^+\). If \(|E_j^+|=0\), the feature is treated as inactive for
metric computation.

\paragraph{Episode Coverage.}
Episode coverage measures how broadly a feature activates across the dataset:
\(c_j=|E_j^+|/|E|\). A high value of \(c_j\) indicates that feature \(j\) appears
across many episodes or tasks, while a low value indicates that the feature is
rare. Importantly, low coverage does not necessarily mean the feature is dead:
it may correspond to a specific object layout, rare failure mode, or memorized
trajectory fragment.

\paragraph{Mean Onset Count.}
An onset is a transition from inactive to active. For each episode \(e\), the
onset count and mean onset count are
\[
\begin{aligned}
o_j^{(e)}
&=
\sum_{t=1}^{T^{(e)}}
\max\!\left(0,s_{t,j}^{(e)}-s_{t-1,j}^{(e)}\right),
\quad s_{0,j}^{(e)}=0, &
\bar{o}_j
&=
\frac{1}{|E_j^+|}\sum_{e\in E_j^+}o_j^{(e)} .
\end{aligned}
\]
This metric measures burstiness. A high \(\bar{o}_j\) means that the feature
turns on repeatedly within active episodes, which is typical of event-locked
features such as repeated contact, grasping, or approach phases. A low
\(\bar{o}_j\) means that the feature activates once and remains active for a
long interval, which is often associated with episode-level or scene-level
memorization.

\paragraph{Relative Run Length.}
The run length measures how long the feature remains active after each onset:
\[
\begin{aligned}
r_j^{(e)}
&=
\frac{\sum_{t=1}^{T^{(e)}}s_{t,j}^{(e)}}{\max(o_j^{(e)},1)}, &
\ell_{r,j}^{(e)}
&=
\frac{r_j^{(e)}}{T^{(e)}}, &
\bar{\ell}_{r,j}
&=
\frac{1}{|E_j^+|}\sum_{e\in E_j^+}\ell_{r,j}^{(e)} .
\end{aligned}
\]
Values near zero indicate short, bursty activations. Values near one indicate
sustained activations over a large fraction of the episode. For VLAs, long
sustained activations often indicate memorized episode structure. For WAMs,
longer run length can also be meaningful when combined with high future
consistency, because predictive features may persist across imagined or planned
future states.

\paragraph{Mean Activation Magnitude.}
Mean activation magnitude measures the typical peak strength of a feature when
it fires:
\[
a_j^{(e)}=\max_{1\leq t\leq T^{(e)}}f_j(x_t^{(e)}), \qquad
\bar{a}_j=\frac{1}{|E_j^+|}\sum_{e\in E_j^+}a_j^{(e)} .
\]
This metric separates features that barely cross the threshold from features
that activate strongly and consistently. We use it together with coverage and
onset statistics, since a rare feature with high peak activation may still be a
meaningful memorized or rare-reactive feature.

\paragraph{Future Consistency Score.}
Future Consistency Score (FCS) measures whether an SAE feature is aligned with a
future latent state. Let \(d_j\) be the decoder direction of SAE feature \(j\),
and let \(y_{t+\Delta}^{(e)}\) be the aligned future latent target at offset
\(\Delta\). For WAM future streams, \(y_{t+\Delta}^{(e)}\) is the realized
current-observation latent at a later hidden timestep. We define
\[
\begin{aligned}
\bar{y}_{t+\Delta}^{(e)}
&=
\frac{y_{t+\Delta}^{(e)}}
{\|y_{t+\Delta}^{(e)}\|_2+\epsilon}, &
q_{t,j}^{(e,\Delta)}
&=
\cos(d_j,\bar{y}_{t+\Delta}^{(e)})
=
\frac{d_j^\top \bar{y}_{t+\Delta}^{(e)}}
{\|d_j\|_2\|\bar{y}_{t+\Delta}^{(e)}\|_2+\epsilon}.
\end{aligned}
\]
FCS is computed as an activation-weighted average:
\[
\mathrm{FCS}_j(\Delta)
=
\frac{
\sum_{e}\sum_{t=1}^{T^{(e)}-\Delta}
f_j(x_t^{(e)})q_{t,j}^{(e,\Delta)}
}{
\sum_{e}\sum_{t=1}^{T^{(e)}-\Delta}
f_j(x_t^{(e)})+\epsilon
},
\qquad
\mathrm{FCS}_j
=
\frac{1}{H}\sum_{\Delta=1}^{H}\mathrm{FCS}_j(\Delta).
\]
A high FCS means that when the feature activates, its decoder direction is
consistent with future latent states. This is especially meaningful for WAM
future streams. For WAMs, FCS is computed on explicit future/imagination streams. For direct VLAs, which do not expose imagined future latents, FCS is computed only as a post-hoc alignment between current/action SAE features and realized future rollout latents. Thus, VLA FCS should be interpreted as temporal correlation with future states, not as evidence of explicit future imagination.

\paragraph{Horizon Stability.}
Horizon Stability (HS) measures whether a feature remains coherent across future
timesteps. For a scalar SAE activation, we use a conditional persistence score:
\[
\mathrm{HS}_j(h)
=
\frac{
\sum_e\sum_{t=1}^{T^{(e)}-h}
s_{t,j}^{(e)}s_{t+h,j}^{(e)}
}{
\sum_e\sum_{t=1}^{T^{(e)}-h}
s_{t,j}^{(e)}+\epsilon
},
\qquad
\mathrm{HS}_j
=
\frac{1}{H}\sum_{h=1}^{H}\mathrm{HS}_j(h).
\]
This score can be interpreted as the probability that a feature remains active
over the next \(H\) timesteps, conditioned on being active now. Low HS indicates
a short event-triggered feature. High HS indicates that the feature persists
over a rollout horizon. For WAMs, high HS together with high FCS suggests a
temporally stable predictive feature. For VLAs, high HS without FCS may instead
indicate sustained memorization or scene-level activation.

\paragraph{Magnitude-Weighted Horizon Stability.}
As an optional variant, we also compute a magnitude-weighted stability score:
\[
\mathrm{HS}^{\mathrm{mag}}_j(h)
=
\frac{
\sum_e\sum_{t=1}^{T^{(e)}-h}
f_j(x_t^{(e)})f_j(x_{t+h}^{(e)})
}{
\sqrt{\sum_e\sum_t f_j(x_t^{(e)})^2}
\sqrt{\sum_e\sum_t f_j(x_{t+h}^{(e)})^2}
+\epsilon
}.
\]
This variant is useful when a feature is active over many timesteps but its
activation strength varies smoothly over the episode. In the main analysis, we
use the binary HS score for interpretability and report the magnitude-weighted
version only as an auxiliary check.

\paragraph{Action Predictiveness.}
Action Predictiveness (AP) measures whether a feature is informative about
future actions. Let \(a_t^{(e)}\) be the executed action or action chunk at
timestep \(t\). For horizon \(H_a\), define the flattened future action segment
as
\[
A_{t,H_a}^{(e)}
=
\mathrm{vec}\!\left(
a_t^{(e)},a_{t+1}^{(e)},\ldots,a_{t+H_a-1}^{(e)}
\right).
\]
For each feature \(j\), we fit a scalar linear predictor from the feature
activation to the future action segment,
\[
\hat{A}_{t,H_a}^{(e)}
=
\beta_{0,j}+\beta_{1,j}f_j(x_t^{(e)}),
\]
and define AP using an \(R^2\) score:
\[
\mathrm{AP}_j
=
1-
\frac{
\sum_{e,t}
\left\|A_{t,H_a}^{(e)}-\hat{A}_{t,H_a}^{(e)}\right\|_2^2
}{
\sum_{e,t}
\left\|A_{t,H_a}^{(e)}-\bar{A}\right\|_2^2+\epsilon
}.
\]
Here \(\bar{A}\) is the mean future action segment. In practice, this score can
be computed with a held-out split or cross-validation to avoid overestimating
predictiveness. High AP indicates that a feature is informative about future
control, not merely about the current visual state.

\paragraph{Alignment Across Different Temporal Resolutions.}
Some models produce one hidden activation per policy query, while others execute
a chunk of multiple low-level actions per query. When hidden timesteps and
action timesteps are not identical, we align them using the policy query index.
If hidden timestep \(t\) corresponds to low-level action interval
\([\ell_t,\ell_{t+1})\), then
\[
A_{t,H_a}^{(e)}
=
\mathrm{vec}\!\left(
a_{\ell_t}^{(e)},a_{\ell_t+1}^{(e)},\ldots,
a_{\ell_t+H_a-1}^{(e)}
\right).
\]
This prevents AP from unfairly favoring models that log actions at a finer
temporal resolution.

\paragraph{Dead Feature Filtering.}
A feature is excluded as dead only when it has no meaningful activation across
episodes. In implementation, we use the strict rule
\[
\max_{e,t}\left|f_j(x_t^{(e)})\right|=0
\quad\Rightarrow\quad
\text{dead feature}.
\]
We do not mark a feature as dead only because it has low coverage. This is
important because low-coverage features may correspond to memorized episodes,
rare task phases, rare object layouts, or failure recovery behaviors.

\paragraph{Metric Vector for Classification.}
For each non-dead feature, we collect
\[
m_j=
\left[
c_j,
\bar{o}_j,
\bar{\ell}_{r,j},
\bar{a}_j,
\mathrm{FCS}_j,
\mathrm{HS}_j,
\mathrm{AP}_j
\right],
\qquad
P(y_j\mid m_j),
\]
where
\[
y_j\in
\{
\mathrm{memorized},
\mathrm{reactive\text{-}general},
\mathrm{predictive\text{-}general}
\}.
\]
Missing values are allowed when a metric is not applicable to a stream. For
example, AP is unavailable if no aligned action trace exists, and FCS is
unavailable if no future latent target exists. In these cases, missing values
are imputed from non-dead features within the same SAE group before classifier
training.

\paragraph{Interpretation.}
The metrics should not be interpreted independently. A low-coverage feature with
high activation magnitude may be memorized rather than inactive. A high-run-length
feature may be memorized if it has low FCS, but predictive if it also has high
FCS and appears in a WAM future stream. Similarly, high AP indicates control
relevance, but not necessarily future prediction. We therefore use these metrics
jointly and treat the resulting labels as weakly supervised probabilistic
annotations rather than hard ground truth.

\subsection{Feature Labelling, Classification, and Verification}
\label{app:feature_labeling}

This appendix describes how SAE features are assigned semantic labels after the
feature-space metrics in Appendix~\ref{app:feature_metrics} have been computed.
The purpose of this procedure is not to obtain perfect ground-truth labels for
every SAE latent. Instead, we use a weakly supervised labelling pipeline to
estimate the distribution of feature types across WAM and VLA latent groups.

\subsubsection{Feature Types and Dead Feature Filtering}
\label{app:feature_types}

For each non-dead SAE feature \(j\), we assign probabilities over three primary
semantic types:
\[
y_j
\in
\{
\mathrm{memorized},
\mathrm{reactive\text{-}general},
\mathrm{predictive\text{-}general}
\}.
\]
These labels describe the role of a feature in the policy representation, not
whether the feature is useful or harmful. A predictive feature is not inherently
better than a reactive feature; the two capture different computational roles.

\textbf{Memorized.} A feature is labeled \emph{memorized} if it activates meaningfully but only in a
small number of episodes, layouts, objects, task variants, failures, or
trajectory fragments. Such features often have low episode coverage and
top-activating examples that share a specific visual or trajectory pattern. They
are not treated as dead if their activations are real and interpretable.

\textbf{Reactive-general.} A feature is labeled \emph{reactive-general} if it responds to the current
observation, robot state, action, or execution phase. These features describe
what is happening now, such as approach, grasp, lift, carry, place, release,
retract, or recovery. They usually show clear phase alignment across multiple
episodes but do not need to activate before the relevant event.

\textbf{Predicitve-general.} A feature is labeled \emph{predictive-general} if it provides evidence about a
future latent state, robot phase, or action. Importantly, a feature is not
predictive merely because it comes from a future stream. We assign this label
only when there is future-oriented evidence, such as high future consistency,
temporal persistence across horizons, activation before the corresponding future
phase, or correlation with future action chunks.

During manual inspection, we also use auxiliary labels for quality control:
\[
\{
\mathrm{rare\text{-}reactive},
\mathrm{candidate\text{-}predictive},
\mathrm{uncertain},
\mathrm{inactive/dead}
\}.
\]
A rare-reactive feature aligns with a phase but appears in too few episodes to
be considered general. A candidate-predictive feature shows future correlation
but lacks sufficient timing evidence for a predictive-general label. An
uncertain feature has nontrivial activation but no reliable interpretation. An
inactive/dead feature has no meaningful activation across episodes.

We use a conservative dead-feature rule because low coverage alone may indicate
memorization rather than inactivity:
$
c_j \approx 0
\;\centernot\Rightarrow\;
\mathrm{inactive/dead}.
$
A feature is excluded only if it has no meaningful activation. In implementation,
the strict dead-feature criterion is
\[
\max_{e,t}
\left|
f_j(x_t^{(e)})
\right|
=
0.
\]
For manual inspection, we also check whether maximum activations are near zero,
whether active timesteps are negligible, and whether top-activating examples are
empty or noisy. If a feature has meaningful top activations, it is inspected as a
possible memorized, rare-reactive, candidate-predictive, or uncertain feature
rather than removed as dead.

\begin{table*}[h]
\centering
\scriptsize
\setlength{\tabcolsep}{3pt}
\renewcommand{\arraystretch}{0.95}
\caption{Manual audit label counts for SAE features. Rows are grouped by model
family. Train features denote labels used for training the feature classifier, while
main-analysis features denote labels retained for the final analysis.}
\label{tab:manual_audit_counts}
\resizebox{\textwidth}{!}{
\begin{tabular}{clrrrrrrrrr}
\toprule
&
Model
& \makecell{\#Total\\original}
& Memorized
& Reactive
& \makecell{Rare\\reactive}
& \makecell{Candidate\\predictive}
& Predictive
& Uncertain
& \makecell{Train\\feats}
& \makecell{Main-analysis\\feats} \\
\midrule
\rowcolor{jointwam}
& Cosmos     & 45 & 15 & 13 & 2 & 11 & 4 & 0  & 32 & 32 \\
\rowcolor{seqwam}
& LingBot-VA & 45 & 15 & 15 & 0 & 11 & 4 & 0  & 32 & 34 \\
\rowcolor{auxwam}
& FastWAM    & 45 & 15 & 16 & 0 & 13 & 1 & 0  & 32 & 32 \\
\rowcolor{auxwam}
\multirow{-4}{*}{\rotatebox[origin=c]{90}{\textsc{WAM}}}
& VLA-JEPA   & 45 & 15 & 10 & 5 &  6 & 9 & 0  & 34 & 34 \\
\midrule
& \(\pi_0\)     & 45 & 15 & 14 & 2 &  0 & 0 & 14 & 28 & 29 \\
& \(\pi_{0.5}\) & 45 & 15 & 14 & 3 &  0 & 0 & 13 & 29 & 29 \\
\multirow{-3}{*}{\rotatebox[origin=c]{90}{\textsc{VLA}}}
& X-VLA         & 45 & 15 & 20 & 0 & 10 & 0 & 0  & 35 & 45 \\
\bottomrule
\end{tabular}}
\end{table*}

\subsubsection{Manual Seed Labelling and Phase Evidence}
\label{app:manual_seed_labeling}

The feature classifier is trained from a small manually inspected seed set. For
each model and SAE group, we initially sample approximately \(45\) features,
covering diverse metric regimes rather than only the most active features. These
include features with high, medium, and low episode coverage, different onset
counts and run lengths, high future consistency, high action predictiveness, and
random active features. This sampling strategy encourages the classifier to see
both easy and ambiguous cases.

For each sampled feature \(j\), we inspect its metric vector
\[
m_j
=
[
c_j,
\bar{o}_j,
\bar{\ell}_{r,j},
\bar{a}_j,
\mathrm{FCS}_j,
\mathrm{HS}_j,
\mathrm{AP}_j
],
\]
together with its top-activating episodes and activation timeline. When rollout
states are available, we also inspect end-effector motion, gripper state, object
motion, reward, and done signals. Each feature is assigned a semantic label, an
optional dominant phase, a confidence score, and a short rationale. We do not
force the final label distribution to be balanced, since some models may
naturally contain more reactive, memorized, or predictive features than others.

Labels are assigned using a simple decision procedure. We first check whether
the feature has meaningful activation; features with no coherent top examples or
near-zero activation are marked inactive/dead, while weak but nontrivial cases
are marked uncertain. If a feature activates only in a few highly specific
episodes, layouts, objects, failures, or trajectory fragments, it is labeled
memorized. If it aligns with the current execution phase, such as approach,
grasp, lift, carry, place, release, or retract, it is labeled reactive-general
when it appears broadly, and rare-reactive when it appears only rarely. If the
feature activates before a later phase, has high future consistency, remains
stable across future horizons, or predicts future action chunks, it is labeled
predictive-general when the evidence is strong and candidate-predictive when the
evidence is suggestive but insufficient. Features without reliable semantic
evidence are labeled uncertain and excluded from classifier training. We visualize the activation of each type of feature for each model as follows.

\begin{figure}[h]
    \centering
    \includegraphics[width=\linewidth]{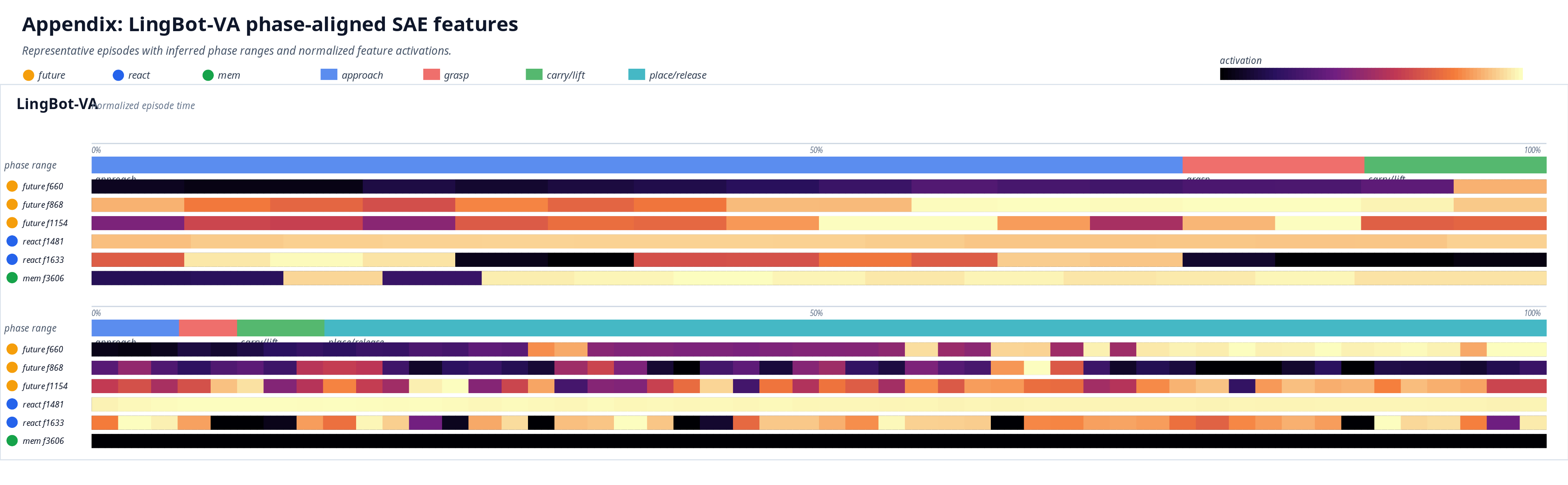}
    \caption{\small Phase-aligned sparse feature activations of Lingbot-VA.}
    \label{fig:hello}
    \vspace{-0.3cm}
\end{figure}

\begin{figure}[h]
    \centering
    \includegraphics[width=\linewidth]{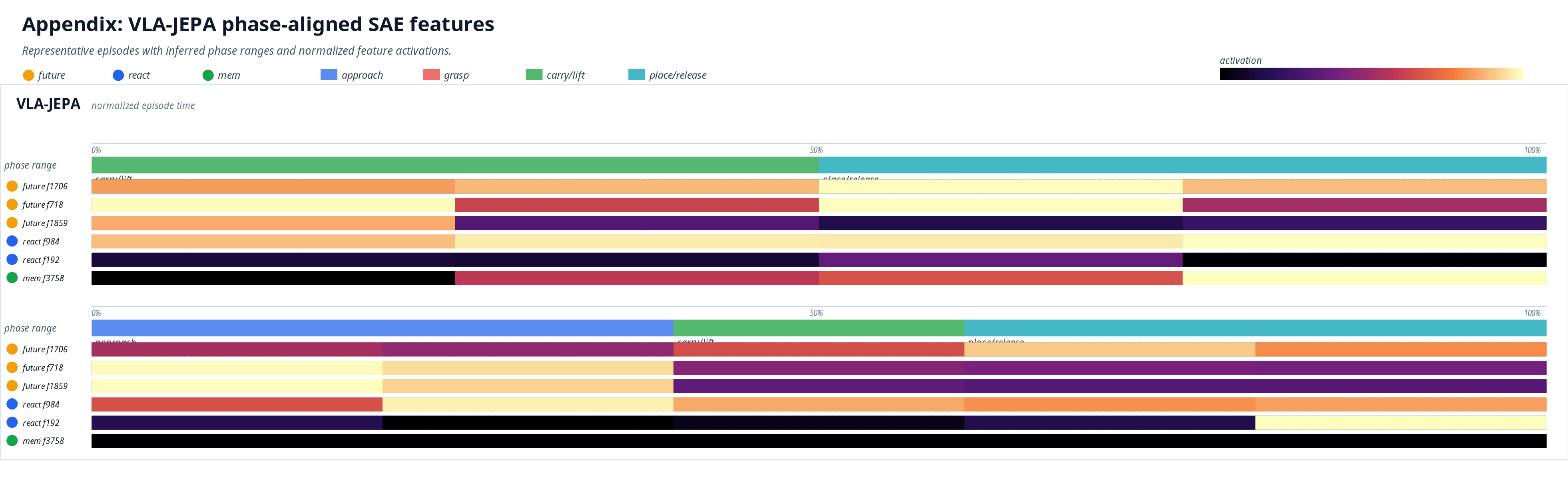}
    \caption{\small Phase-aligned sparse feature activations of VLAJEPA.}
    \label{fig:hello}
    \vspace{-0.3cm}
\end{figure}

\begin{figure}[h]
    \centering
    \includegraphics[width=\linewidth]{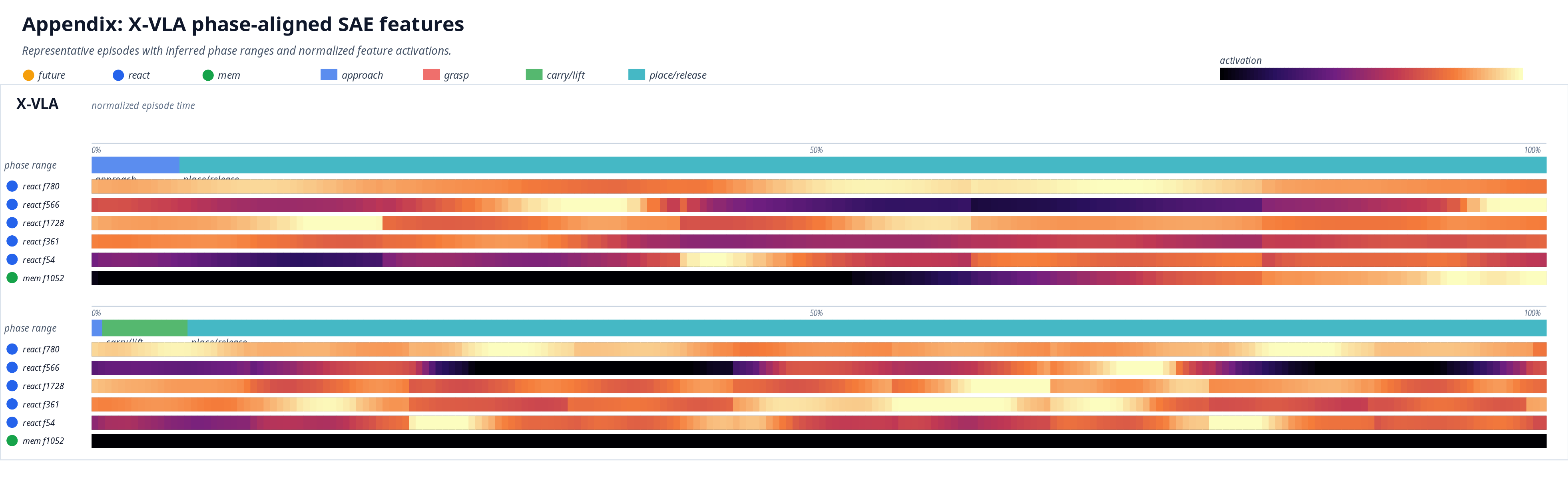}
    \caption{\small Phase-aligned sparse feature activations of XVLA. }
    \label{fig:hello}
    \vspace{-0.3cm}
\end{figure}

\begin{figure}[h]
    \centering
    \includegraphics[width=\linewidth]{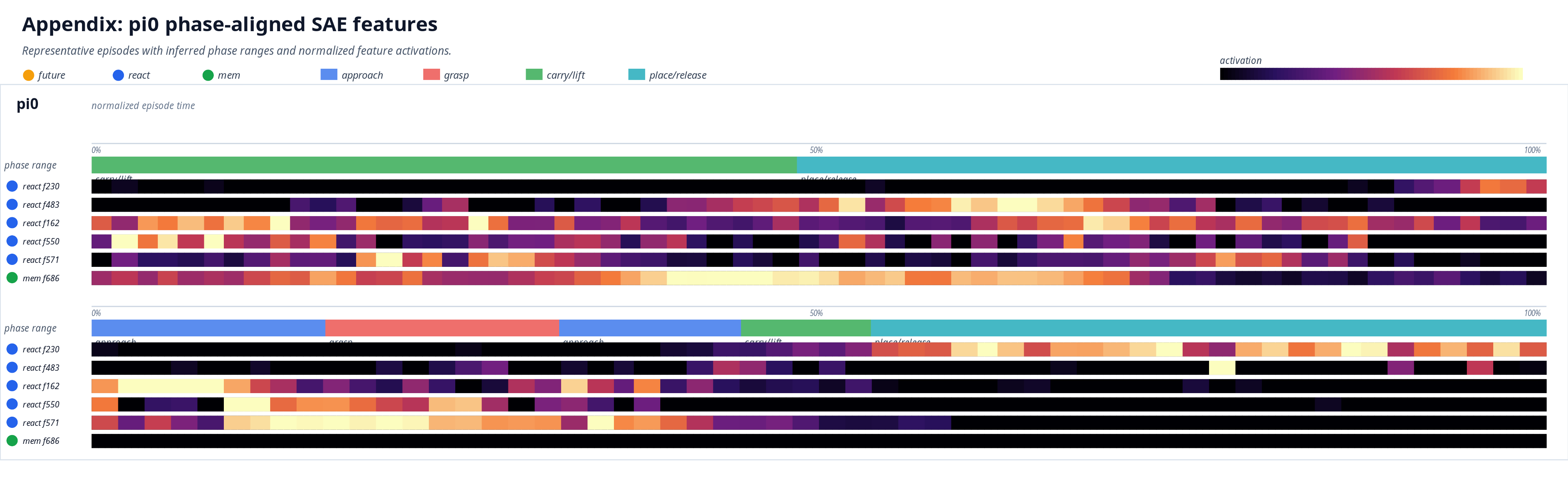}
    \caption{\small Phase-aligned sparse feature activations of $\pi_0$.}
    \label{fig:hello}
    \vspace{-0.3cm}
\end{figure}

\begin{figure}[h]
    \centering
    \includegraphics[width=\linewidth]{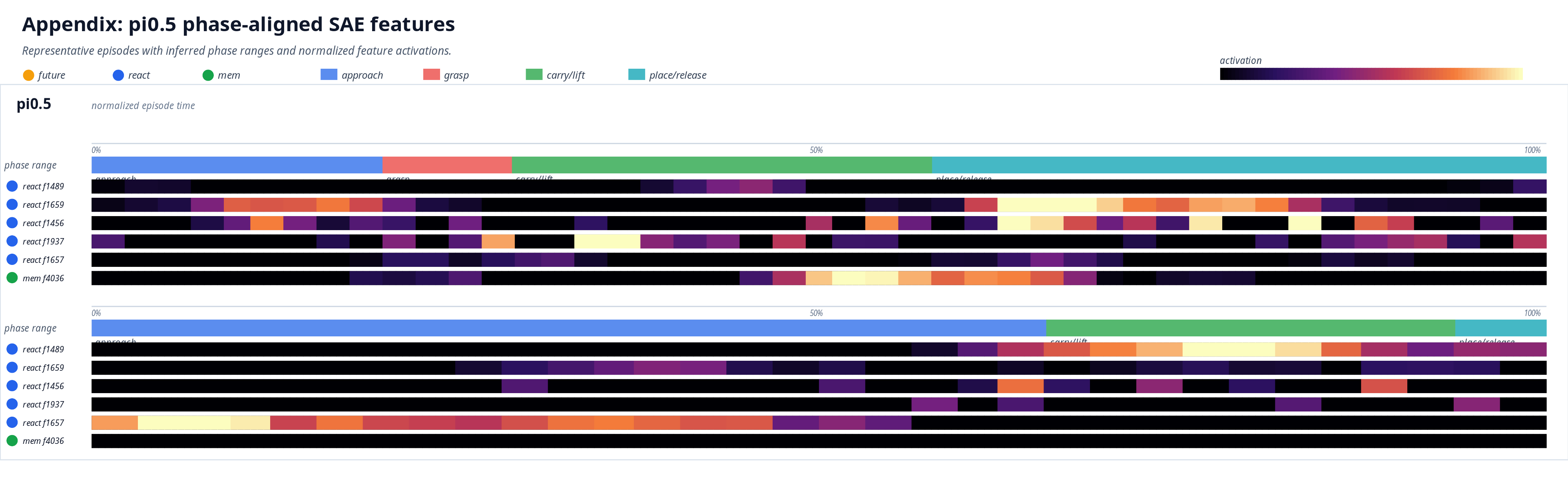}
    \caption{\small Phase-aligned sparse feature activations of $\pi_{0.5}$. }
    \label{fig:hello}
    \vspace{-0.3cm}
\end{figure}

To support phase interpretation, we estimate a dominant phase for each feature.
Let
\[
\Phi
=
\{
\mathrm{approach},
\mathrm{grasp},
\mathrm{lift/carry},
\mathrm{release/place}
\}.
\]
For each timestep, we estimate a soft phase distribution
\(P(\phi_t=\phi\mid \mathrm{rollout})\) from rollout states when available, and
from a weaker normalized-time proxy otherwise. The feature-level phase
probability is computed by activation-weighted aggregation:
\[
P(\phi\mid j)
=
\frac{
\sum_e \sum_t
f_j(x_t^{(e)})
P(\phi_t^{(e)}=\phi\mid \mathrm{rollout})
}{
\sum_e \sum_t f_j(x_t^{(e)}) + \epsilon
}.
\]
The predicted phase is
$
\hat{\phi}_j
=
\arg\max_{\phi\in\Phi}
P(\phi\mid j).
$
This phase label is used only as auxiliary evidence and is not treated as ground
truth unless verified by manual inspection.

\begin{table*}[h]
\centering
\scriptsize
\setlength{\tabcolsep}{3.0pt}
\renewcommand{\arraystretch}{0.95}
\caption{\small \textbf{Aggregate SAE feature metrics by dataset and model.}
Numeric metrics are averaged over active SAE features. Dead percentage is computed
over all learned SAE features. Row colors group WAM model families.}
\label{tab:sae_metric_aggregates}
\resizebox{\linewidth}{!}{
\begin{tabular}{clrrrrrrrrrr}
\toprule
&
Model
& \makecell{\#Feat.}
& \makecell{\#Active}
& \makecell{Dead\\(\%)}
& Cov.
& \makecell{Run\\len.}
& Onsets
& \makecell{Future\\cons.}
& \makecell{Action\\pred.}
& \makecell{Horizon\\stab.}
& \makecell{Active\\frac.} \\
\midrule
\multicolumn{12}{c}{\textbf{LIBERO}} \\
\midrule
\rowcolor{jointwam}
& Cosmos & 28,672 & 21,431 & 25.3 & 0.012 & 0.108 & 1.033 & 0.019 & 0.004 & 0.095 & 0.002 \\
\rowcolor{seqwam}
& LingBot-VA & 14,336 & 8,508 & 40.7 & 0.932 & 0.468 & 2.237 & -0.006 & 0.016 & 0.711 & 0.634 \\
\rowcolor{auxwam}
& FastWAM & 11,264 & 3,533 & 68.6 & 0.342 & 0.195 & 1.388 & 0.035 & 0.010 & 23.282 & 0.124 \\
\rowcolor{auxwam}
\multirow{-4}{*}{\rotatebox[origin=c]{90}{\textsc{WAM}}}
& VLA-JEPA & 9,216 & 6,568 & 28.7 & 0.108 & 0.367 & 1.056 & 0.031 & 0.008 & 0.286 & 0.055 \\
\midrule
& $\pi_0$ & 5,120 & 2,291 & 55.3 & 0.136 & 0.584 & 1.536 & 0.000 & 0.014 & 0.693 & 0.064 \\
& $\pi_{0.5}$ & 10,240 & 5,330 & 47.9 & 0.125 & 0.120 & 1.298 & 0.011 & 0.004 & 0.272 & 0.037 \\
\multirow{-3}{*}{\rotatebox[origin=c]{90}{\textsc{VLA}}}
& X-VLA & 4,096 & 2,555 & 37.6 & 0.170 & 0.203 & 1.362 & 0.009 & 0.004 & 27.550 & 0.050 \\
\midrule
\multicolumn{12}{c}{\textbf{RoboTwin}} \\
\midrule
\rowcolor{seqwam}
& LingBot-VA & 10,240 & 5,120 & 50.0 & 0.134 & 0.266 & 1.433 & 0.062 & 0.001 & 0.314 & 0.052 \\
\rowcolor{auxwam}
& FastWAM & 11,264 & 5,831 & 48.2 & 0.190 & 0.335 & 1.494 & 0.032 & 0.004 & 39.527 & 0.094 \\
\midrule
& $\pi_{0.5}$ & 5,120 & 2,860 & 44.1 & 0.112 & 0.242 & 1.373 & 0.000 & 0.003 & 0.281 & 0.051 \\
\bottomrule
\end{tabular}}
\end{table*}

\subsubsection{Weakly Supervised Type--Phase Classifier}
\label{app:feature_type_classifier}

After manual seed labelling, we train a weakly supervised probabilistic
classifier to assign feature-type probabilities to each non-dead SAE feature.
Each feature \(j\) is represented by the metric vector
\[
m_j
=
[
c_j,
\bar{o}_j,
\bar{\ell}_{r,j},
\bar{a}_j,
\mathrm{FCS}_j,
\mathrm{HS}_j,
\mathrm{AP}_j
].
\]
Features are then standardized using statistics from the seed set. The SAE feature metrics statistics are shown in Table~\ref{tab:sae_metric_aggregates}. We use a multinomial logistic classifier over the standardized SAE metric vector and trained with class-balanced weights so that rare labels,
especially predictive-general features, are not ignored:
\[
P(y_j=k\mid m_j)
=
\frac{
\exp(b_k + w_k^\top \tilde{m}_j)
}{
\sum_{k'}
\exp(b_{k'} + w_{k'}^\top \tilde{m}_j)
}.
\]
\[
k \in
\{
\mathrm{memorized},
\mathrm{reactive\text{-}general},
\mathrm{predictive\text{-}general}
\},
\]

For LIBERO, the fitted standardized coefficients for
\((\mathrm{memorized}, \mathrm{reactive}, \mathrm{predictive})\) are
\[
\beta_c=(-0.860,\;0.549,\;0.310),\quad
\beta_o=(-0.220,\;0.272,\;-0.052),
\]
\[
\beta_{\ell_r}=(-0.327,\;-0.428,\;0.755),\quad
\beta_a=(0.043,\;-0.089,\;0.046),
\]
\[
\beta_{\mathrm{FCS}}=(-0.469,\;-0.428,\;0.897),\quad
\beta_{\mathrm{HS}}=(-0.138,\;0.077,\;0.061),
\]
\[
\beta_{\mathrm{AP}}=(-0.832,\;1.047,\;-0.215),
\]
with intercepts
\[
\beta_0=(0.153,\;0.955,\;-1.108).
\]
This classifier achieves \(90.5\%\) cross-validation accuracy and macro-F1
\(0.859\) on \(222\) labeled LIBERO examples.

For RoboTwin, the fitted standardized coefficients are
\[
\beta_c=(-1.138,\;1.051,\;0.088),\quad
\beta_o=(-0.159,\;0.004,\;0.155),
\]
\[
\beta_{\ell_r}=(-0.278,\;-0.737,\;1.015),\quad
\beta_a=(0.453,\;-0.003,\;-0.450),
\]
\[
\beta_{\mathrm{FCS}}=(-0.787,\;0.048,\;0.739),\quad
\beta_{\mathrm{HS}}=(0.105,\;0.315,\;-0.420),
\]
\[
\beta_{\mathrm{AP}}=(-0.523,\;0.833,\;-0.311),
\]
with intercepts
\[
\beta_0=(0.389,\;0.591,\;-0.980).
\]
This classifier achieves \(93.8\%\) cross-validation accuracy and macro-F1
\(0.895\) on \(96\) labeled RoboTwin examples.

\section{Extensive Result}
\subsection{Additional RoboTwin Result}
\label{app:robotwin_audit}

RoboTwin action-space metrics require benchmark-local interpretation. Unlike
LIBERO, RoboTwin logs absolute qpos targets for two arms. Thus, acceleration,
jerk, and boundary jumps measure changes in commanded joint targets rather than
changes in end-effector delta commands. These metrics are still useful within
RoboTwin, but should not be averaged with LIBERO metrics.

\paragraph{Success-conditioned smoothness.}
Table~\ref{tab:robotwin_success_cond} separates successful and failed episodes.
The low aggregate smoothness of \(\pi_{0.5}\) is explained by stalled
failures: 95.3\% of its failed episodes are low-motion episodes. FastWAM shows a
similar but weaker pattern, while Lingbot-VA has far fewer low-motion failures.
This supports the interpretation that smooth RoboTwin qpos trajectories do not
necessarily imply better task completion.

\begin{table*}[h]
\centering
\scriptsize
\setlength{\tabcolsep}{5pt}
\renewcommand{\arraystretch}{0.95}
\caption{Success-conditioned RoboTwin smoothness.}
\label{tab:robotwin_success_cond}
\resizebox{\textwidth}{!}{
\begin{tabular}{llrrrrrr}
\toprule
Model & Outcome & Delta \(\downarrow\) & Accel \(\downarrow\)
& Jerk \(\downarrow\) & Boundary \(\downarrow\)
& B./non-B. \(\downarrow\) & Low-motion eps. \(\downarrow\) \\
\midrule
\(\pi_{0.5}\) & Success & 0.0576 & 0.0159 & 0.0237 & \textbf{0.0102} & 0.281 & 25.1 \\
\(\pi_{0.5}\) & Failure & \textbf{0.0336} & \textbf{0.0135} & \textbf{0.0211} & 0.0110 & 0.507 & 95.3 \\
Lingbot-VA    & Success & 0.0295 & 0.0173 & 0.0311 & 0.0631 & 1.125 & \textbf{25.0} \\
Lingbot-VA    & Failure & 0.0346 & 0.0328 & 0.0599 & 0.0736 & 1.110 & 32.9 \\
FastWAM       & Success & 0.0600 & 0.0208 & 0.0339 & 0.0381 & 0.932 & \textbf{25.0} \\
FastWAM       & Failure & 0.0312 & 0.0216 & 0.0375 & 0.0295 & 1.359 & 99.8 \\
\bottomrule
\end{tabular}}
\end{table*}

\paragraph{RoboTwin object diagnostics.}
RoboTwin object metrics are heuristic because task objects and reference objects
are mapped from task names rather than following the LIBERO target-object schema.
Table~\ref{tab:robotwin_object} should therefore be read as a benchmark-local
diagnostic. Lingbot-VA improves over \(\pi_{0.5}\) in completion-related metrics,
with higher transport and place rates. FastWAM shows the most stable object
interaction: it has the best reference progress, far fewer reversals, shorter
target-object paths, higher directness, and the lowest wrong-object disturbance.
This suggests that, although Lingbot-VA achieves stronger completion than
\(\pi_{0.5}\), FastWAM produces more direct and less distracting object-level
behavior on RoboTwin.

\begin{table*}[h]
\centering
\scriptsize
\setlength{\tabcolsep}{3pt}
\renewcommand{\arraystretch}{0.95}
\caption{\small \textbf{RoboTwin2.0 object interaction diagnostics.}
The table reports the same object-level metrics as LIBERO. Progress metrics are
computed only on episodes with mapped reference objects, so they should be read
as benchmark-local diagnostics.}
\label{tab:robotwin_object}
\resizebox{\textwidth}{!}{
\begin{tabular}{llrrrrrrrrrrr}
\toprule
Family & Model
& \makecell{Prog.\\\(\uparrow\)}
& \makecell{Prog.\\AUC \(\uparrow\)}
& \makecell{Rev.\\\(\downarrow\)}
& \makecell{Target\\path \(\downarrow\)}
& \makecell{Direct.\\\(\uparrow\)}
& \makecell{Wrong\\motion \(\downarrow\)}
& \makecell{Wrong\\path \(\downarrow\)}
& \makecell{Grasp\\(\%) \(\uparrow\)}
& \makecell{Lift\\(\%) \(\uparrow\)}
& \makecell{Transport\\(\%) \(\uparrow\)}
& \makecell{Place\\(\%) \(\uparrow\)} \\
\midrule
VLA & \(\pi_{0.5}\)
& -0.375 & -1.332 & 106.2 & 0.896 & 0.398
& 0.005 & 0.010 & 74.5 & 73.7 & 56.3 & 39.0 \\

\rowcolor{seqwam}
WAM & Lingbot-VA
& -0.113 & -0.451 & 79.4 & 0.924 & 0.413
& 0.093 & 0.106 & 81.7 & 81.7 & 85.1 & 59.6 \\

\rowcolor{auxwam}
WAM & FastWAM
& -0.065 & -0.239 & 4.1 & 0.465 & 0.568
& 0.002 & 0.004 & 74.2 & 74.2 & 71.2 & 53.7 \\
\bottomrule
\end{tabular}}
\end{table*}

Table~\ref{tab:robotwin_runtime} shows a similar runtime pattern on RoboTwin:
direct VLA inference remains the fastest, while WAM-style models introduce
additional deployment cost due to future prediction or rollout-style reasoning.
Among the WAMs, FastWAM is much more efficient than LingBot-VA, suggesting that
future-aware policies can be made substantially cheaper depending on how the
future model is integrated.

\begin{table}[h]
\centering
\scriptsize
\setlength{\tabcolsep}{3pt}
\renewcommand{\arraystretch}{0.92}
\caption{\small Runtime and deployment cost on RoboTwin.}
\label{tab:robotwin_runtime}
\resizebox{0.56\linewidth}{!}{
\begin{tabular}{lrrrrr}
\toprule
Model 
& \makecell{p50\\(ms) \(\downarrow\)} 
& \makecell{p95\\(ms) \(\downarrow\)}
& \makecell{Chunks/s\\\(\uparrow\)}
& \makecell{Eff. Hz\\\(\uparrow\)}
& \makecell{GPU\\(MB)} \\
\midrule
LingBot-VA   & 6444 & 6871 & 0.163 &  4.700 & 32,031 \\
FastWAM      &  687 &  695 & 1.456 & 41.900 & 34,152\\
\midrule
\(\pi_{0.5}\) & 159 & 178 & 6.289 & 62.893 & 10,341 \\
\bottomrule
\end{tabular}}
\end{table}



\paragraph{Chunk-alignment audit.}
For RoboTwin, \(\pi_{0.5}\) and FastWAM action chunks align with executed actions
under the common evaluator. Lingbot-VA, however, stores chunks with shape
\((Q,16,2,16)\), and simple flattening does not reconstruct the executed 16-D
action stream. Therefore, Lingbot-VA RoboTwin chunk-space quantities should be
treated as representation/logging diagnostics rather than directly comparable
policy-quality metrics.

\paragraph{Protocol interpretation.}
The apparent RoboTwin smoothness advantage of \(\pi_{0.5}\) is not a competence
reversal. \(\pi_{0.5}\) uses horizon 32 with query gap 32, consuming the full
chunk before the next query. FastWAM uses horizon 32 with query gap 24 and
therefore switches plans before the old chunk is fully consumed. This early
replanning increases qpos boundary sensitivity. Combined with the high
low-motion failure rate of \(\pi_{0.5}\), the RoboTwin result suggests that
smooth qpos trajectories can correspond to stalled failures, while WAM policies
complete more tasks but incur larger replanning discontinuities during active
manipulation.

\subsection{Additional detail for Feature-space Analysis}
\label{app:feat_detail}
\subsubsection{Detail Analysis}

\paragraph{Combined interpretation across feature, action, and object spaces.}
Taken together, the feature-space, action-space, and object-space results suggest
that the advantage of WAMs is not simply that they achieve higher success, but
that their architectures change how manipulation is internally organized. In
LIBERO, WAM-family policies generally produce smoother executed actions, with
lower acceleration, jerk, boundary discontinuity, and low-motion failure than
direct VLA baselines. The object diagnostics further show that this smoother
control often corresponds to more object-oriented behavior: WAMs tend to make
stronger target progress and reduce unintended distractor motion. The SAE results
provide a complementary explanation for this trend. Models with explicit or
latent future-modeling components expose non-zero predictive feature structure,
whereas direct VLAs are dominated by reactive features. Thus, the behavioral
difference is consistent with a representational difference: WAMs appear to
encode not only the current observation--instruction--action mapping, but also
information about how the scene is expected to evolve.

\textbf{Lingbot-VA.} This interpretation is clearest for the sequential WAM, LingBot-VA. Its current
representation contains the largest predictive fraction among all models, while
its future representation remains highly active and contains both reactive and
predictive structure. This matches the imagine-then-act design: before the
future state is realized, the current representation must encode what should
happen next; after the future branch is constructed, many features become
reactive with respect to the imagined state itself. This architectural separation
also helps explain its rollout behavior. LingBot-VA achieves very smooth actions,
low boundary discontinuity, strong target progress, high progress AUC, and very
few progress reversals. In other words, its explicit future branch seems to
support temporally coherent manipulation, where the policy can maintain a stable
notion of the next object-centric phase instead of repeatedly reacting to the
current frame.

\textbf{Cosmos}, as a joint WAM, shows a different but still informative pattern. It has
many active features in current, future, and action spaces, but most are labeled
as memorized rather than predictive. This does not necessarily mean that future
information is absent. Instead, it suggests that the future signal is entangled
inside a shared generative trajectory representation, rather than isolated into
clean sparse predictive variables. This interpretation is consistent with its
rollout behavior: Cosmos achieves high success, low action irregularity, and the
lowest wrong-object disturbance in LIBERO. Its object behavior is therefore
highly selective, but the SAE does not recover this as a large set of simple
predictive features. A plausible explanation is that joint generative WAMs encode
future evolution as high-dimensional video-trajectory templates, which are useful
for avoiding distractor interaction but harder to decompose into separate
future-state features.

\textbf{FastWAM.} The auxiliary WAMs illustrate why the WAM label alone is insufficient. FastWAM
has smooth action-space behavior, but its current and future SAE spaces are very
sparse, with high dead-feature rates and only a small predictive fraction. Its
action space is much more active, suggesting that the future-prediction signal is
compressed into the final policy rather than preserved as an explicit
inference-time representation. This helps explain why FastWAM can look strong in
low-level motion metrics while being weaker in object selectivity: its
wrong-object path and wrong-object motion are close to the VLA range. In other
words, action smoothness alone does not guarantee object-aware manipulation. If
future modeling is used mainly as a training regularizer, the policy may inherit
some motion stability but lose part of the explicit spatial reasoning that helps
separate the target object from distractors.

\textbf{VLA-JEPA} behaves differently from FastWAM and is closer to an intermediate case.
Its future space retains a moderate predictive fraction and its object
disturbance is much lower than FastWAM and the direct VLA baselines. This is
consistent with a JEPA-style latent prediction objective: the model learns
future-sensitive structure in representation space, even if it does not perform
full pixel-level or rollout-level imagination at inference time. However, its
action representation is extremely small in terms of active features, so the high
predictive percentage in the action row should not be over-interpreted. The more
important pattern is that VLA-JEPA preserves some future-oriented latent
structure and correspondingly shows cleaner distractor behavior than FastWAM.
This supports our broader argument that latent future prediction can be useful,
but only if the learned future representation remains accessible to action
selection rather than being reduced to a weak auxiliary signal.

\textbf{VLA.} The direct VLA baselines provide the contrast case. For \(\pi_0\) and
\(\pi_{0.5}\), predictive SAE features are absent in both current and action
spaces, and their active features are mainly reactive. This matches their
architecture: they condition on the current visual-language context and map it
directly to actions without an explicit future-state branch. Their rollout
behavior also follows this pattern. Although \(\pi_{0.5}\) reaches high success
on LIBERO, its object metrics show higher progress reversals and substantially
larger wrong-object disturbance than the strongest WAMs. X-VLA is somewhat
different because its current representation is mostly memorized, suggesting
that it may encode task- or demonstration-level templates, but its action space
is still reactive and it does not expose predictive features. Overall, the VLA
results suggest that strong current-state grounding and large-scale training can
produce competent manipulation, but they do not automatically yield explicit
future-oriented structure or the same degree of object-level selectivity.

These results therefore support the discussion in two ways. First, they suggest
that future-oriented representations are behaviorally meaningful: models that
retain clearer future structure tend to show smoother execution, stronger target
progress, and cleaner object interaction. Second, they show that the form of
future modeling matters. Sequential WAMs expose future information most clearly
but are computationally expensive; joint WAMs can produce strong object
selectivity but encode future information in a more entangled way; auxiliary WAMs
are more efficient but may compress or weaken the future signal at inference
time. Thus, the design goal should not be to remove imagination entirely, but to
make it lightweight and usable for action selection. A promising direction is to
preserve explicit or latent future prediction during inference, while avoiding
the full runtime cost of sequential video rollout. This would keep the main
benefit observed in our analysis---future-aware, object-selective manipulation---
without making WAMs impractical for deployment.

\subsection{Result on RoboTwin}

Table~\ref{tab:robotwin_model_space_feature_summary} reports the SAE feature classification result on RoboTwin; overall, the report presents the same trend as that of LIBERO.

\begin{table}[h]
\centering
\scriptsize
\setlength{\tabcolsep}{3.5pt}
\renewcommand{\arraystretch}{0.95}
\caption{\small \textbf{RoboTwin per-model SAE feature classification by representation space.}
\#Active Feat. denotes the number of active non-dead SAE features. Dead percentage is
computed over all learned SAE features. Percentages for memorized, reactive, and
predictive features are computed over non-dead features with joint weak labels.}
\label{tab:robotwin_model_space_feature_summary}
\resizebox{0.6\textwidth}{!}{
\begin{tabular}{cllrrrrr}
\toprule
&
Model
& Space
& \makecell{\#Active \\ Feat.}
& \makecell{Dead\\(\%)}
& \makecell{\textbf{Mem.}\\(\%)}
& \makecell{\textbf{React.}\\(\%)}
& \makecell{\textbf{Pred.}\\(\%)} \\
\midrule

\rowcolor{auxwam}
&         & Current & 2,660 & 35.1 & 71.0 & 26.5 & 2.5 \\
\rowcolor{auxwam}
& FastWAM & Future  & 440   & 89.3 & 19.8 & 76.3 & 3.9 \\
\rowcolor{auxwam}
&         & Action  & 2,743 & 10.7 & 62.3 & 35.2 & 2.5 \\

\noalign{\vskip 2pt}

\rowcolor{seqwam}
&            & Current & 1,929 & 52.9 & 70.3 & 11.5 & 18.2 \\
\rowcolor{seqwam}
& LingBot-VA & Future  & 2,138 & 47.8 & 62.7 & 10.2 & 27.1 \\
\rowcolor{seqwam}
\multirow{-6}{*}{\rotatebox[origin=c]{90}{\textsc{WAM}}}
&            & Action  & 1,061 & 48.2 & 59.0 & 23.3 & 17.7 \\

\midrule

& $\pi_{0.5}$ & Current & 2,694 & 34.2 & 95.5 & 4.5  & 0.0 \\
\multirow{-2}{*}{\rotatebox[origin=c]{90}{\textsc{VLA}}}
&             & Action  & 168   & 83.6 & 19.6 & 80.4 & 0.0 \\

\bottomrule
\end{tabular}}
\end{table}

\subsection{Additional SAE-Behavior Probe Results}
\label{app:sae_behavior_probe}

\begin{figure*}[h]
    \centering
    \includegraphics[width=\textwidth]{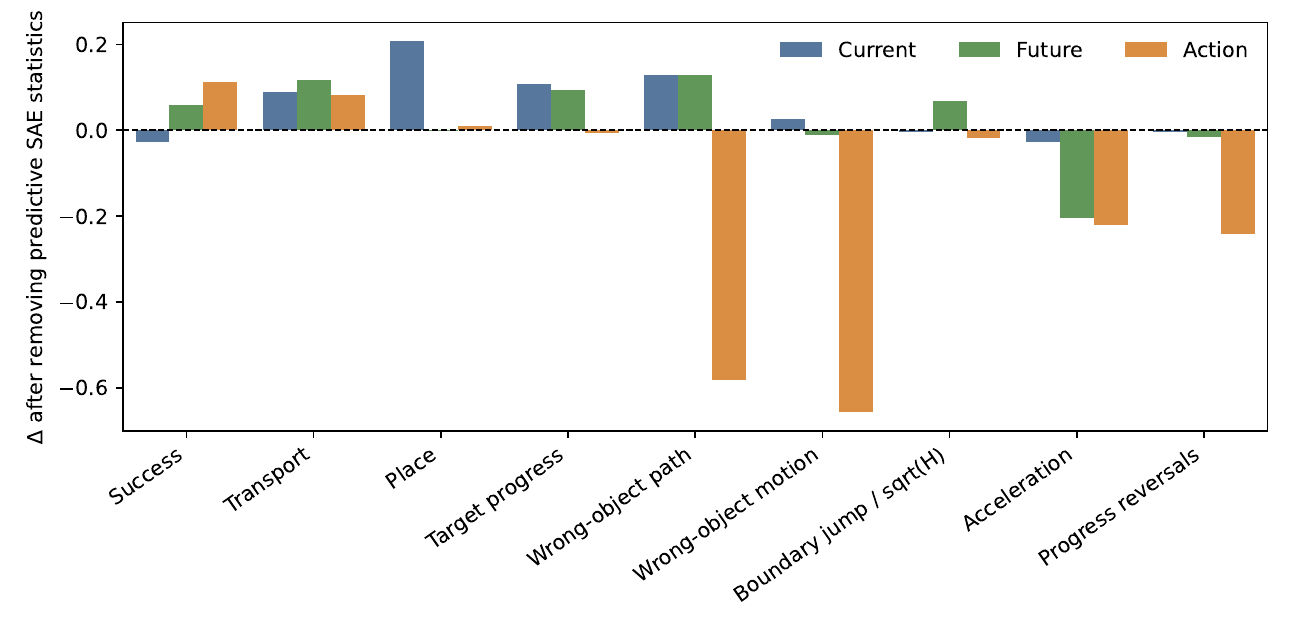}
    \caption{\small \textbf{SAE-behavior probe deltas.}
    Each bar reports the change in probe performance after removing predictive
    SAE statistics, computed as $\Delta = \text{score}_{\text{all}} -
    \text{score}_{\text{w/o pred.}}$. Positive values indicate that predictive
    SAE statistics improve behavioral prediction.}
    \label{fig:appendix_sae_behavior_delta}
\end{figure*}

To test whether the predictive SAE features identified in the representation
analysis are behaviorally informative, we aggregate SAE statistics at the
episode level and use them to predict rollout outcomes. For each episode, we
compute activation mass by feature type, active feature ratios, activation-
weighted future consistency score (FCS), activation-weighted horizon stability,
activation-weighted action predictiveness, and top predictive-feature activation
mass. We then train lightweight probes with task-level GroupKFold splits to
avoid leakage across episodes from the same task. Binary outcomes are evaluated
with AUC, while continuous outcomes are evaluated with $R^2$. We compare probes
using all SAE statistics against probes where predictive SAE statistics are
removed; therefore, a positive $\Delta$ indicates that predictive SAE statistics
add behavioral information. Figure~\ref{fig:appendix_sae_behavior_delta}
visualizes these deltas across representation spaces, and
Table~\ref{tab:sae_behavior_probe_appendix_delta} reports the corresponding
values. These results should be interpreted as associative evidence rather than
as a causal intervention on the policy.

\begin{table*}[h]
\centering
\scriptsize
\setlength{\tabcolsep}{4pt}
\renewcommand{\arraystretch}{0.95}
\caption{\small \textbf{Score change after removing predictive SAE statistics.}
Positive values indicate that predictive SAE statistics improve behavioral
prediction. The mixed signs for low-level metrics show that predictive features
are more strongly associated with task/object-level behavior than with all
smoothness diagnostics.}
\label{tab:sae_behavior_probe_appendix_delta}
\begin{tabular}{llccc}
\toprule
Target & Score & Current $\Delta$ & Future $\Delta$ & Action $\Delta$ \\
\midrule
Success & AUC & -0.028 & +0.060 & \textbf{+0.112} \\
Transport & AUC & +0.090 & \textbf{+0.116} & +0.081 \\
Place & AUC & \textbf{+0.208} & -0.001 & +0.010 \\
Target progress & $R^2$ & \textbf{+0.108} & +0.095 & -0.005 \\
Wrong-object path & $R^2$ & +0.128 & \textbf{+0.129} & -0.582 \\
Wrong-object motion & $R^2$ & \textbf{+0.026} & -0.012 & -0.657 \\
Boundary jump$/\sqrt{H}$ & $R^2$ & -0.004 & \textbf{+0.067} & -0.018 \\
Acceleration & $R^2$ & -0.027 & -0.204 & -0.222 \\
Progress reversals & $R^2$ & -0.003 & -0.015 & -0.242 \\
\bottomrule
\end{tabular}
\end{table*}

Table~\ref{tab:sae_behavior_probe_appendix_delta} summarizes the same ablation
across representation spaces. Predictive SAE statistics provide the clearest
gain for completion-related and object-level metrics, including success,
transport, target progress, and wrong-object path. In contrast, low-level
smoothness metrics such as acceleration and progress reversals show mixed or
negative deltas. This suggests that predictive SAE features are more closely
associated with task and object-level manipulation quality than with all forms
of low-level action regularity. The latter can also depend on the action decoder,
chunking protocol, and controller interface. Establishing a causal role for
these features would require direct feature intervention during policy rollout.

\subsection{SAE Ablation and Null-Control Validation}
\label{app:sae_ablation}

To make the SAE-based feature analysis credible, we need to rule out two simple failure modes. First, an SAE could produce features that look interpretable but do not faithfully reconstruct the original hidden activations, or it could rely on a small number of active latents while most dictionary elements are dead. Second, the future-consistency score (FCS) could be an artifact of the decoder geometry, task identity, or episode ordering rather than evidence of structured future-related information. This appendix therefore serves as a sanity check for the feature-space analysis rather than as a new model comparison. We keep only the compact SAE-health/null-normalized table and the real-vs-null FCS figure because they directly answer these two questions. The full SAE-health grid is mostly diagnostic, the label-threshold sensitivity results belong to weak-label robustness, and the future-vs-past contrast is omitted because it was not run for the full grid.

\begin{table}[h]
\centering
\scriptsize
\setlength{\tabcolsep}{4pt}
\renewcommand{\arraystretch}{0.95}
\caption{\small \textbf{Compact SAE ablation and null-normalized future consistency on FastWAM LIBERO-10.}
NMSE denotes normalized reconstruction error. Dead percentage is computed over all learned SAE features.
The default and \textsc{aux-off} runs are numerically identical in these completed runs, indicating that the
auxiliary dead-latent loss was effectively inactive. $z_{\mathrm{FCS}}$ normalizes the real FCS against the
available null controls.}
\label{tab:sae_ablation_compact}
\resizebox{\linewidth}{!}{%
\begin{tabular}{llrrrrr}
\toprule
Run / stream & Seed & Setting & NMSE $\downarrow$ & Dead (\%) $\downarrow$ & Active feat. & $z_{\mathrm{FCS}}$ $\uparrow$ \\
\midrule
Current, representative & 101 & Default / \textsc{aux-off} & 0.000504 & 0.269 & 4085 & 0.58 \\
Future, representative & 101 & Default / \textsc{aux-off} & 0.001994 & 0.000 & 4096 & 6.59 \\
Future, aux-active stress & 202 & Default / \textsc{aux-off} & 0.002512 & 0.000 & 4096 & 9.60 \\
\bottomrule
\end{tabular}%
}
\end{table}

Table~\ref{tab:sae_ablation_compact} shows that the representative FastWAM LIBERO-10 SAEs are well behaved: they reconstruct the analyzed hidden streams with low NMSE and have almost no dead features. The current stream has only (0.269\%) dead features, while both future-stream runs use all 4096 learned latents. Intuitively, this means that the sparse dictionaries are not merely fitting noise with a collapsed or mostly inactive basis. Disabling the auxiliary dead-latent loss gives identical NMSE, dead-feature rate, active-feature count, FCS, and AP in these completed runs. Thus, we do not claim that AuxK is unnecessary in general; rather, the result shows that the FastWAM feature statistics reported here are not driven by an active auxiliary-loss effect.

\begin{figure}[h]
\centering
\includegraphics[width=\linewidth]{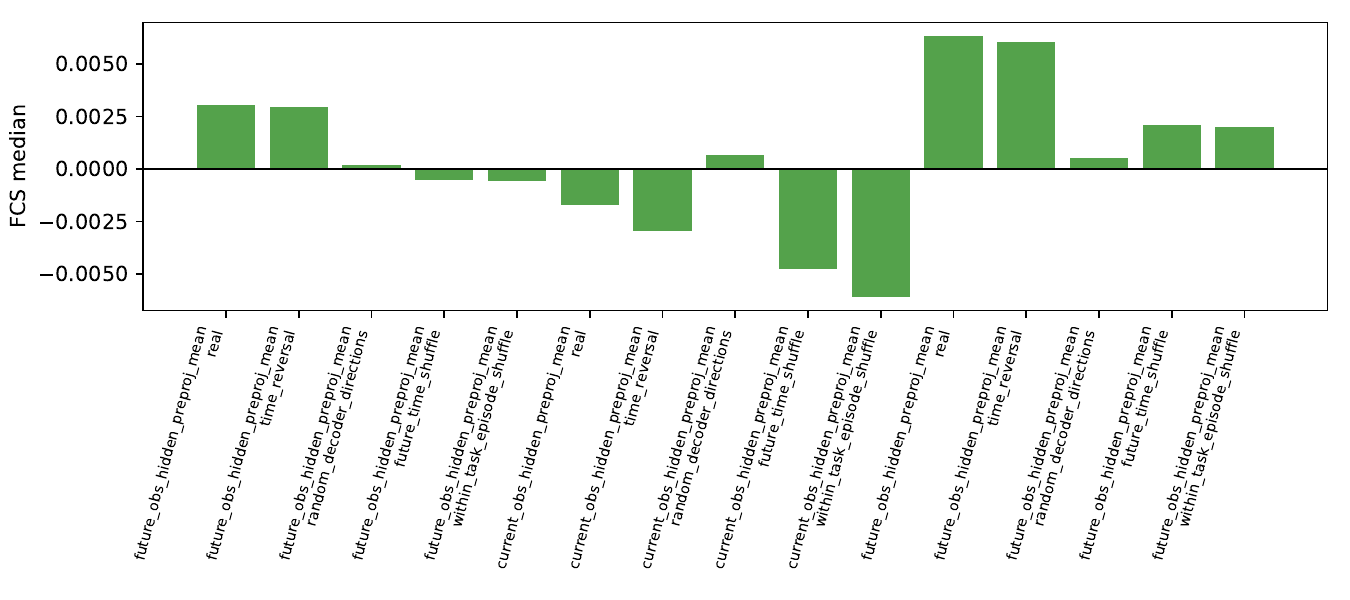}
\caption{\small \textbf{Real-vs-null future consistency.}
Real future-stream features retain higher median FCS than random decoder directions and temporal/episode shuffles, while the current stream does not show the same positive future-aligned structure.}
\label{fig:sae_real_vs_null_fcs}
\end{figure}

Figure~\ref{fig:sae_real_vs_null_fcs} then asks whether the future-consistency signal still appears when the feature--future relationship is deliberately broken. If FCS only came from arbitrary decoder directions or from task-level correlations, then random decoder directions, future-time shuffling, or within-task episode shuffling should give similar values to the real pairing. Instead, the future stream keeps a positive and clearly higher FCS than these null controls, and the same pattern appears in the seed-202 stress run. The current stream, by contrast, stays near zero or negative and does not show the same future-aligned structure. This supports the interpretation that future-stream SAE features capture structured variation associated with future states. However, because time reversal does not substantially reduce FCS, and because the future-vs-past contrast was not run for the full grid, these controls do not prove a strict temporal arrow or causal influence on behavior. We therefore treat SAE labels as probabilistic feature annotations and interpret SAE--behavior relationships as associative evidence.

Table~\ref{tab:sae_ablation_compact} shows that the representative FastWAM LIBERO-10 SAEs reconstruct the analyzed hidden streams with low error and almost no dead features. The current stream has (0.269

Figure~\ref{fig:sae_real_vs_null_fcs} further tests whether future consistency is preserved under null controls. In the representative future stream, real FCS remains positive and is substantially above random decoder directions and future/episode shuffles; the same pattern holds in the seed-202 stress run. By contrast, the current stream has real FCS near zero or negative and does not exhibit the same future-aligned signal. However, time reversal does not substantially reduce FCS, and the future-vs-past contrast was not run for the full grid. These controls therefore support the presence of structured future-associated variation in future-stream SAE features, but they do not establish a strict temporal arrow or causal influence on robot behavior. We consequently use SAE labels as probabilistic annotations and interpret SAE--behavior relationships as associative evidence.

\subsection{Multi-seed training ablation}
\begin{figure*}[h]
    \centering

    \includegraphics[
        width=0.84\textwidth,
        trim=0 16pt 0 0,
        clip
    ]{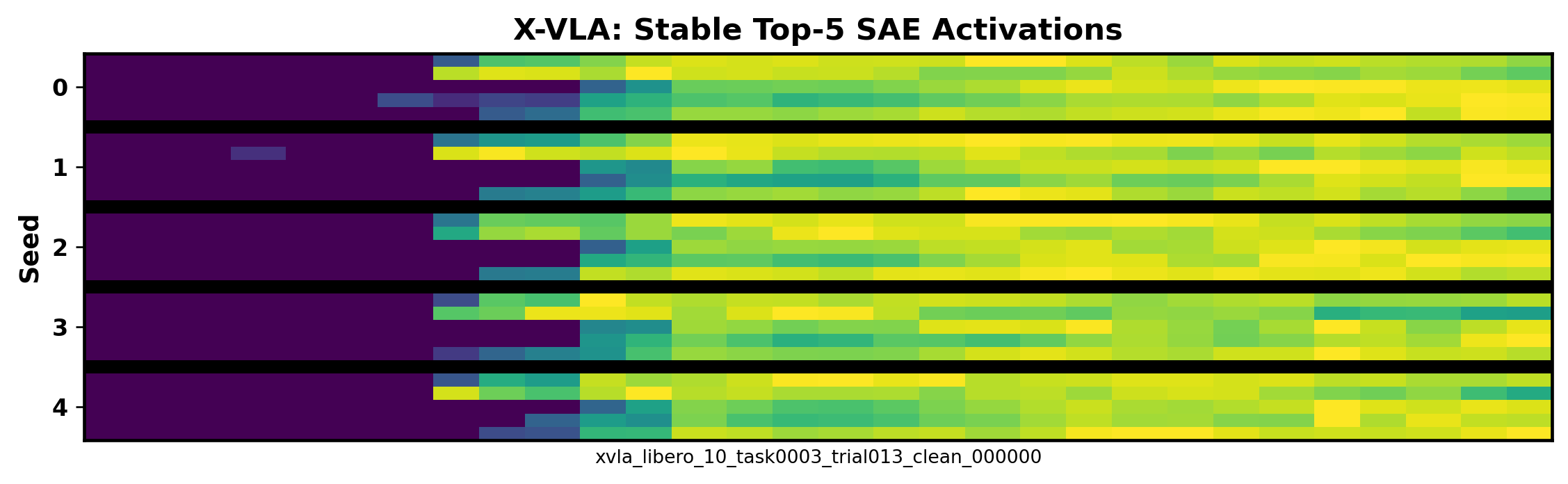}
    \vspace{-1mm}
    \centerline{\small (a) X-VLA}

    \vspace{2mm}

    \includegraphics[
        width=0.84\textwidth,
        trim=0 16pt 0 0,
        clip
    ]{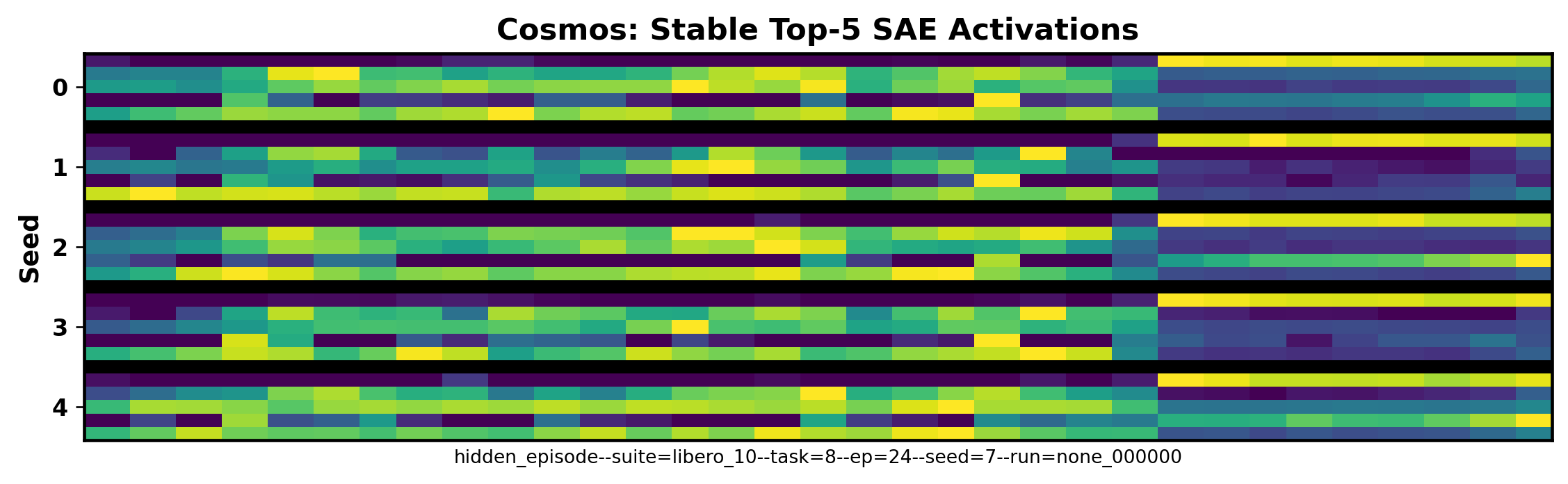}
    \vspace{-1mm}
    \centerline{\small (b) Cosmos}

    \caption{\small \textbf{Multi-seed SAE stability ablation.}
    We train SAEs from 5 independent random seeds on the same activation data and
    visualize the stable top-5 features for a fixed rollout. Activations are
    normalized for visualization and ordered by cross-seed consistency. Both
    X-VLA and Cosmos exhibit recurring temporal activation patterns across
    independent SAE initializations, suggesting that the selected features
    reflect structure in the underlying policy representations rather than
    seed-specific SAE artifacts.}
    \label{fig:multiseed_sae_stability}
\end{figure*}

To verify that the SAE features used in our analysis are not artifacts of a
particular random initialization, we train multiple SAEs with 5 independent seeds
on the same activation data and compare the most stable top features for a fixed
rollout. Figure~\ref{fig:multiseed_sae_stability} shows that both X-VLA and
Cosmos recovers temporally consistent activation patterns across seeds. In
X-VLA, the strongest features are mostly inactive at the beginning and then
activate persistently after the manipulation phase begins. Cosmos also shows
stable cross-seed structure, although its activations are more heterogeneous,
with several features alternating between early/middle activity and late-stage
activation. These shared temporal patterns across independent SAE trainings
suggest that the selected features reflect structure in the underlying policy
representations rather than seed-specific SAE artifacts.

\end{document}